\documentclass[twoside,11pt]{article}

\usepackage[preprint]{jmlr2e}

\jmlrheading{1}{2021}{1-48}{6/21}{10/00}{meila00a}{Andrea Bassich, Francesco Foglino, Matteo Leonetti, and Daniel Kudenko}

\ShortHeadings{Curriculum Learning with a Progression Function}{Bassich et al.}
\firstpageno{1}

\usepackage{dblfloatfix}
\usepackage{algorithm}
\usepackage{algorithmic}
\usepackage[switch]{lineno}

\begin{document}

\title{Curriculum Learning with a Progression Function}

\author{\name Andrea Bassich \email ab1770@york.ac.uk \\
       \addr Department of Computer Science\\
       University of York
       \AND
       \name Francesco Foglino \email francesco@phantasma.global \\
       \addr Phantasma Labs GmbH
       \AND
       \name Matteo Leonetti \email matteo.leonetti@kcl.ac.uk \\
       \addr Department of Informatics\\
       King's College London
       \AND
       \name Daniel Kudenko \email kudenko@l3s.de \\
       \addr L3S research center\\
       Leibniz Universität Hannover}

\editor{n.a.}

\maketitle

\begin{abstract}
Curriculum Learning for Reinforcement Learning is an increasingly popular technique that involves training an agent on a sequence of intermediate tasks, called a Curriculum, to increase the agent's performance and learning speed. This paper introduces a novel paradigm for curriculum generation based on progression and mapping functions. While progression functions specify the complexity of the environment at any given time, mapping functions generate environments of a specific complexity.
Different progression functions are introduced, including an autonomous online task progression based on the agent's performance. 
Our approach's benefits and wide applicability are shown by empirically comparing its performance to two state-of-the-art Curriculum Learning algorithms on six domains.
\end{abstract}

\begin{keywords}
  curriculum learning, reinforcement learning, progression functions
\end{keywords}

\section{Introduction}

Curriculum Learning for Reinforcement Learning has been an increasingly active area of research; its core principle is to train an agent on a sequence of intermediate tasks, called a Curriculum, to increase the agent's performance and learning speed. Previous work has focused on gradually modifying the agent's experience within a given task~\citep{florensa2017,pmlr-v80-riedmiller18a} or scheduling sequences of different and increasingly complex tasks~\citep{narvekar2019learning,foglino2019optimization}.

We introduce a framework for Curriculum Learning that encompasses both paradigms, centred around the concept of task complexity and its progression as the agent becomes increasingly competent. The framework enables ample flexibility as the tasks can be selected from an infinite set and, most significantly, the task difficulty can be modified at each time step. Furthermore, this framework is learning-algorithm agnostic, as it focuses on modifying the environment according to the agent's ability and does not need access to the agent's internal state.
The framework is based on two components: a \emph{progression function} calculating the appropriate complexity of the task for the agent at any given time, and a \emph{mapping function} modifying the environment according to the required complexity. The progression function encompasses task selection and sequencing, while the mapping function is responsible for task generation. Generation, selection and sequencing are the central components of Curriculum Learning~\citep{narvekar2016source}, seamlessly integrated into the proposed framework.

Unlike previous work, our method can directly leverage simple high-level domain knowledge in generating a Curriculum for the agent. Even in extremely complex domains, defining how the environment's complexity could be modified is often straightforward. For example, if an agent's objective is to learn how to drive a car on the road, decreasing the friction between the wheels and the ground, simulating slippery conditions, will result in the environment becoming more complex. Likewise, initialising the agent further away from the goal in a maze navigation domain will result in the environment being harder to solve. Our algorithm can leverage this simple domain knowledge to automatically generate a Curriculum that adjusts the complexity of the environment according to the agent's ability.



In the remainder of the paper, we first lay out the mathematical framework behind our approach. We then introduce two classes of progression functions, \emph{fixed} and \emph{adaptive}, based on whether the curriculum is predetermined or generated online. Furthermore, we propose an adaptive progression for the ordering and selection of the tasks in the curriculum. Lastly, we demonstrate the effectiveness of the curriculum learning algorithms derived from the proposed framework on six experimental domains and compare them with state-of-the-art methods for Task-level curricula and experience ordering within the same task.

\section{Background}

In this section, we give a brief overview of Reinforcement Learning to then discuss Curriculum Learning.

\subsection{Reinforcement Learning}

Reinforcement Learning (RL) is a popular field in Machine Learning used to solve tasks requiring the agent to take actions in an environment and adjust its behaviour according to a reward function. More formally, tasks are modelled as episodic Markov Decision Processes (MDP). An MDP is a tuple $\langle S,A,p,r,\gamma \rangle$, where $S$ is the set of states, $A$ is the set of actions, $p : S \times A  \times S \rightarrow [0,1]$ is the transition function, $r: S \times A \rightarrow \mathbb{R}$ is the reward function and $\gamma \in [0, 1]$ is the discount factor. Episodic tasks have \emph{absorbing} states, which cannot be left and from which the agent only receives a reward of $0$. By definition, MDPs also need to satisfy the Markov property: a state carries all the information needed to define any possible future state given a sequence of actions, regardless of the history of the task.

To learn, the agent interacts with the environment at each time-step $t$ by first observing the current state of the environment and then acting according to a policy $\pi: S \times A \rightarrow [0,1]$. After acting in the environment, the agent is provided with the next observation and a reward, a cycle that will repeat until the agent reaches an absorbing state and the episode ends. The agent aims to find the \textit{optimal} policy $\pi^*$ that maximizes the expected \emph{return}. The return from the initial state is defined as $G_0 = \sum_{t=0}^{t_M} \gamma^t r(S_t, A_t)$, where $t_M$ is the maximum length of the episode. 

\subsection{Curriculum Learning}

Whenever we, as humans, need to learn a complex task, our learning is usually organised in a specific order: starting from simple concepts and progressing onto more complex ones as our knowledge increases. The order in which concepts are learnt is often referred to as Curriculum, and it is not limited to human learning. In fact, Curricula are also used to teach animals very complex tasks, like in \cite{skinner1951teach} where a pigeon was trained to recognize the suits in a pack of cards. The idea of applying a Curriculum to Supervised learning dates back to \cite{elman1993learning} where it is suggested that initially training a neural network on a more simple subset might be advantageous. This idea was then formalised by \cite{bengio2009curriculum} and named Curriculum Learning.


When applying this technique to Reinforcement Learning we first define $\mathcal{M}$ as a possibly infinite set of MDPs that are candidate tasks for the curriculum, and $m_f \in \mathcal{M}$, as the \emph{final} task. The final task is the task the designer wants the agent to learn more efficiently through the curriculum. We define a curriculum as a sequence of tasks in $\mathcal{M}$:



\noindent
{\bf Definition} {\it 
Given a set of tasks $\mathcal{M}$,
a \emph{curriculum} over $\mathcal{M}$ of length $l$ is a sequence
of tasks $c=\langle m_{1},m_{2},$ $\ldots,$ $m_{l}\rangle$ where each $m_{i}\in\mathcal{M}$.

} 

\noindent
The tasks in the curriculum are called the \emph{intermediate} tasks.
The Curriculum aims to achieve the highest performance possible within the number of learning steps allocated. Another aim of using a Curriculum is to achieve this performance as quickly as possible. The specific performance metric optimised depends on the Curriculum Learning algorithm: for example HTS-CR \citep{foglino2019curriculum} aims to optimise cumulative reward. In order to keep our framework as flexible as possible we allow the optimisation of both environment specific metrics (such as success rate in a navigation domain) and more traditional metrics such as cumulative reward. This will be expanded in section \ref{sec:adaprog}.




\section{Related Work} \label{sec:related_work}

Learning through a curriculum consists of modifying the agent's experience so that each learning step is the most beneficial. 
The general principle underlying curriculum learning spans a vast gamut of methods: from reordering the generated transition samples within a given task, such as in Prioritized Experience Replay \citep{schaulICLR16} to learning a Curriculum MDP \citep{narvekarIJCAI17} over the space of all possible policies. 

While one may construct a curriculum on the transition samples within a single MDP (as mentioned above for Prioritized Experience Replay), this work belongs to the category of task-level curriculum methods \citep{narvekar2020curriculum}, where the agent is presented with a sequence of different MDPs over time. On the other hand, our method does not impose any restrictions on the differences between the MDPs in a curriculum.

Curriculum Learning for Reinforcement Learning has been formalized by \cite{narvekar2016source}, who also provided heuristics for creating the intermediate tasks.
Most Task-level methods that impose no restrictions on the MDPs focus on selecting and sequencing the tasks, assuming that a set of candidate intermediate tasks exists. In all existing methods, such a set is finite, whilst in our framework, we allow for a potentially infinite set of candidate tasks.

Previous work has considered the automatic generation of a graph of tasks based on a measure of transfer potential~\citep{leonetti17}, later extended to an object-oriented framework that enables random task generation~\citep{daSilva2018}. In both these works, the curriculum is generated before execution (with online adjustments in the latter case), whilst in our framework, the curriculum can be entirely generated online. \cite{narvekar2019learning} introduce an online curriculum generation method based on the concept of Curriculum MDP (CMDP). A CMDP is defined according to the agent's knowledge representation (in particular according to the parameter vector of the policy); for any current policy of the agent, a policy over the CMDP returns the next task to train on. The framework has the theoretical guarantees of MDPs and suffers from the same practical limitations imposed by non-linear function approximation, where convergence to a globally optimal policy cannot be guaranteed.

Teacher-Student Curriculum Learning \citep{matiisen2017teacher} is based on a Partially Observable MDP formulation. The teacher component selects tasks for the student whilst learning, favouring tasks in which the student is making the most progress or that the student appears to have forgotten. The aim is for the agent to be able to solve all the tasks in the curriculum. Conversely, in our problem definition, the intermediate tasks are stepping stones towards the final task, and our agent aims at minimizing the time required to learn the optimal policy for the final task.

Another category of methods is designed with specific restrictions over the set of MDPs in the curriculum, such as when only the reward function changes~\citep{pmlr-v80-riedmiller18a}. In this category, Reverse Curriculum Generation (RCG) ~\citep{florensa2017} gradually moves the start state away from the goal using a uniform sampling of random actions. The agent's performance from the given start states is then assessed to determine which states should be used to generate starts in the algorithm's next iteration. Unlike this algorithm, our method can modify other parameters besides the starting state, and the way our approach forms a curriculum does not include a sampling of random actions. However, the algorithm requires the same limited access to the environment and agent as our approach, and the pace of the curriculum created by this algorithm is akin to our progression functions. Moreover this algorithm is able to generate a Curriculum where the source task changes after every episode, like our algorithm. For these reasons, this approach was selected as one of the baselines to benchmark our method.

Self-Paced Deep Reinforcement Learning \citep{klink2020deep} uses the Contextual RL framework to let the agent modify the context distribution during the training. This algorithm assumes that the target context distribution is known and seeks to balance the current reward with the proximity to the target distribution. Unlike this algorithm, our method does not use the contextual Reinforcement Learning framework. This framework uses a contextual variable $c \in R^n$ to define a Markov Decision Process in a given domain. The algorithm defines its lower and upper bounds, and performs its optimisation within these bounds, however, it does not allow for a subset of contextual variables between the bounds to be excluded from the search. This would imply that assuming that the contextual variable $c$ influences the initial position of the agent's model in a navigation domain ($c \in R^2$, $R^3$ or $R^{15}$ in our navigation testing domains), it would not be possible to exclude invalid contextual variables corresponding with states where the agent cannot be initialised (inside walls, on top of obstacles or in the case of the "Ant" model in positions where its legs clip through the floor). This makes this algorithm not applicable in four of our six evaluation domains (Figure \ref{fig:visual}). Given that this algorithm belongs to the class of Curriculum Learning methods that modify the agent’s experience within a given task, we considered Reverse Curriculum Generation as a better suited representative of this class of algorithms for our comparison.


The formulation of sequencing as a combinatorial optimization problem~\citep{foglino2019optimization} over the intermediate tasks lends itself to globally optimal sequencing algorithms. One such algorithm is Heuristic Task Sequencing for Cumulative Return~\citep{foglino2019curriculum} (HTS-CR), a complete anytime algorithm, converging to the optimal curriculum of a maximum length. This algorithm was chosen as a baseline specifically because of this guarantee; in fact, comparing against this algorithm removes the need to compare against other sequencing algorithms using a predefined set of intermediate tasks.

\section{Mapping and Progression Functions} \label{sec:map_and_prog}


As previously introduced, our curriculum learning framework comprises two elements: a \emph{progression} function, and a \emph{mapping} function. The role of the Progression Function is to specify the appropriate complexity value, between 0 and 1, at each time step. The Mapping Function takes as input the complexity value and generates an MDP with the desired complexity. 

The Point Mass Maze domain will be used as a running example for the rest of the paper. The agent is required to navigate through a G-shaped maze, shown in Figure \ref{fig:visual} (top centre), to reach the goal in the top right corner of the maze. In this domain, the complexity of the environment is modified by changing the starting position of the agent in the maze, where the easiest possible task (of complexity 0) has a starting position right next to the goal, while the hardest task (of complexity 1)  has a starting position as far as possible, along the maze, from the goal.

In this section we also introduce two families of progression functions: fixed and adaptive. The former produces a predetermined curriculum before execution, while the latter results in a dynamic curriculum, adjusted based on the agent's online performance. In addition to implementing task selection and sequencing, the progression function determines how long the agent should train on each task, which is an open problem in other Curriculum Learning methods. In most existing methods, the agent learns intermediate tasks until convergence; however, it has been suggested that this is unnecessary~\citep{narvekar2020curriculum}. Progression functions address this problem by providing an implicit way to determine when to stop training on a task and advance to the next. 

\subsection{Framework} \label{sec:framework}

We define a progression function $\Pi$ as follows:

\begin{equation}
\label{eq:4.1}
\Pi : \mathbb{N}^+ \times  P \rightarrow [0,1]
\end{equation}

\noindent where $c_t = \Pi(t, P)$, is the \emph{complexity value} at time $t$, and $P$ is a set of parameters specific to each progression function. The value $c_t$ reflects the difficulty of the MDP where the agent trains at time $t$. The MDP corresponding to $c_t = 1$ is the final task, and the smaller the value of $c_t$, the easier the corresponding task is.


To generate tasks of a specific complexity,  we introduce the Mapping function, which maps a specific value of $c_t$ to a Markov Decision Process $M$ created from the set  $\mathcal{M}$:
\begin{equation}
\label{eq:4.2}
\Phi_{D}: [0, 1] \rightarrow M
\end{equation}
The co-domain of $\Phi_{D}$ contains the set of all candidate intermediate tasks. The Mapping Function encodes some high-level knowledge about the domain, such as, in the Point Mass Maze domain: the furthest from the goal, the more complex the task. This knowledge is then used to generate a task of the desired complexity, mapping a complexity value (e.g. 0.3) to the MDP with the corresponding set of parameters (e.g. distance = 0.3 * maximum distance). Assuming that the learning time of an MDP is defined as the number of actions needed to converge to the optimal policy from a randomly initialised policy; an ideal mapping function generates the MDP with the lowest learning time when the required complexity is 0 and the MDP with the highest learning time when the required complexity is 1. Moreover, given two MDPs, the difference between their complexity should be proportional to the difference in their learning time.





At any change of the value of $c_t$ according to the progression function, a new intermediate task is added to the agent's curriculum. Given a Mapping function $\Phi$, to the value of $c_t$ at time $t$ corresponds a new intermediate task:

\begin{equation}
\label{eq:4.3}
M_i = \Phi(c_t)
\end{equation}

\noindent
where $i$ is the number of updates to the value of $c_t$ since the start of the progression. Specifying the way $c_t$ changes over time through a progression function selects which tasks are added to the curriculum and in which order, resulting in the curriculum:

\begin{equation}
 \label{eq:4.4}
C = \langle M_0, ..., M_i\rangle 
\end{equation}

\noindent
Algorithm \ref{algo:clpf} shows how our approach is integrated with the agent's learning, illustrating when the environment's complexity is calculated after every action, and the MDP is also updated at most after every action.




\begin{algorithm}[t]
\caption{CL with a Progression Function}
\textbf{Input}: Progression function $\Pi$, mapping function $\Phi$\\
\begin{algorithmic}[1] 
\STATE $t$ := 1
\STATE $c_0$ := 0
\STATE $current\_c$ := 0
\STATE $i$ := 0
\STATE $M_0$ := $\Phi(0)$
\WHILE{learning}
\STATE $s_t$, $r_t$ := result of agent's action in $M_i$
\STATE $c_t$ := $\Pi(t, P)$
\IF {$c_t \neq current\_c$}
\STATE $current\_c$ := $c_t$
\STATE $i$ := $i + 1$ 
\STATE $M_i$ := $\Phi(c_t)$
\ENDIF
\STATE $t := t + 1$
\ENDWHILE
\end{algorithmic}
\label{algo:clpf}

\end{algorithm}

\subsection{Progression Functions}

As mentioned above, progression functions determine how the complexity of the environment changes throughout the training, and in doing so, they generate a Curriculum. In this section, we introduce two types of progression functions: Fixed and Adaptive. Fixed progression functions correspond to pre-defined curricula, whereas Adaptive progression functions correspond to online curriculum generation.


\subsubsection{Fixed progression}


The first class of progression functions used in this paper is called \emph{Fixed} progression, and it includes progression functions that define the curriculum before execution.


The most basic example of one such progression function is the \emph{linear} progression. The only parameter in this function is the time step in which the progression  ends, $t_e$. The equation of this progression function is as follows:

\begin{equation}
\label{eq:4.5}
\Pi_l(t, t_e) := min(\frac{t}{t_e}, 1)
\end{equation}

\noindent When using this progression function, as the name suggests, the complexity of the environment will increase linearly until reaching a value of 1 at $t = t_e$. 


We also introduce a second progression function, the \emph{exponential} progression, with Equation:




 
\begin{equation}
\label{eq:4.6}
\Pi_e(t, \{t_e, s\}) = \frac{\left( e^{\frac{t}{t_e}} \right)^\alpha -1 }{e^{\alpha} -1}
\end{equation}
\begin{equation}
\alpha := \frac{1}{s}
\end{equation}

It includes a parameter to indicate when the progression should end, $t_e$, and a parameter $s$, which influences the progression's slope. A positive value of $s$ results in the progression being steeper at the start and slower towards the end of the training. On the other hand, a negative value of $s$ results in the initial progression being slower. Figure \ref{fig:prog} shows how different values of $s$ affect the progression.

In Equation \ref{eq:4.6} the numerator is responsible for the progression, starting at 0 when t = 0 and increasing as $t$ increases. On the other hand, the denominator serves to guarantee that the value of $c_t$ is equal to 1 as $t = t_e$.



One property of the exponential progression that should be highlighted is that it converges to a linear progression as $s$ tends to infinity (proven in appendix A):

\begin{equation}
\label{eq:linexpeq}
\lim_{s \to \infty} \Pi_e(t, \{t_e, s\}) = \Pi_l(t, t_e)
\end{equation}


The advantages of using a function from the Fixed Progression class is that it is easier to implement than adaptive functions while still retaining the benefits of using a Curriculum. However, the agent will learn differently each time, and using the same progression function does not exploit our framework to its full potential.





\subsubsection{Adaptive progression} \label{sec:adaprog}

\begin{figure*}[!t]
\centering
   \begin{minipage}{0.32\textwidth}
   \centering
     \includegraphics[width=\textwidth]{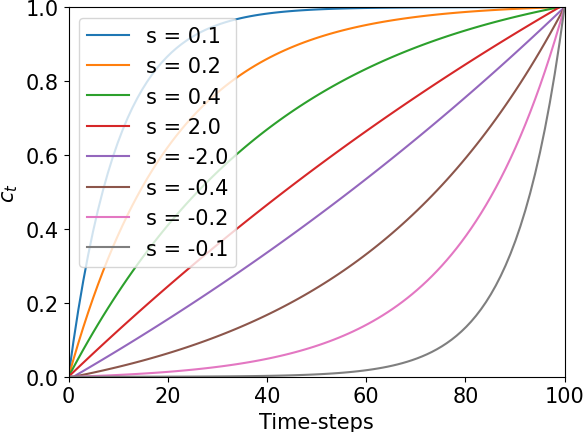}
   \end{minipage}
   \begin{minipage}{0.32\textwidth}
     \centering
     \includegraphics[width=\textwidth]{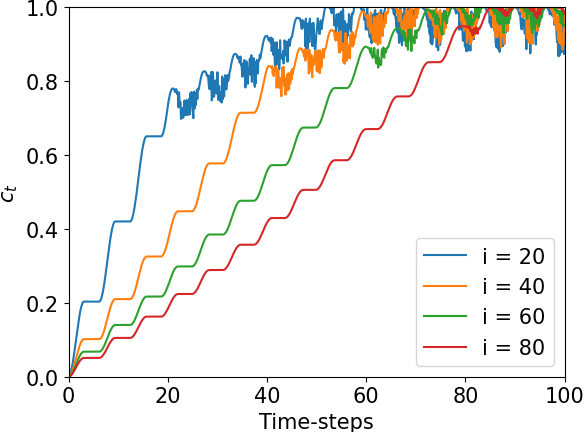}
   \end{minipage}
   \begin{minipage}{0.32\textwidth}
     \centering
     \includegraphics[width=\textwidth]{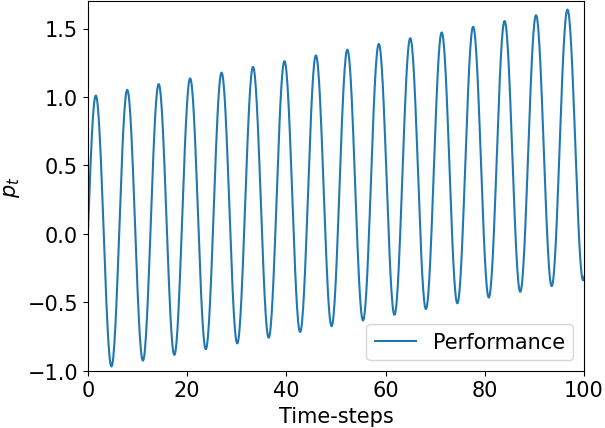}
   \end{minipage}
   
   \caption{Example of different Progression functions. The left figure shows the effect of changing the value of $s$ in an exponential progression. The central figure shows the effect of changing the magnitude of the interval, $i$, on a Friction-based progression, where the performance used for the progression is shown in the figure on the right.} \label{fig:prog}
\end{figure*}

The second class of progression functions introduced in this paper comprises all the functions generated online and that adapt to the agent's performance to create a Curriculum.

Various environment-specific metrics can be used to utilise the agent's performance in progression functions, such as whether the agent reached the goal in the Point Mass Maze domain; more traditional metrics can also be chosen, such as the return. We, therefore, define the \emph{performance function} $p_t \rightarrow R$ as the metric used to evaluate the agent's performance in a given MDP.

\begin{figure}[!t]
\centering
   \begin{minipage}{0.35\textwidth}
     \centering
     \includegraphics[width=\textwidth]{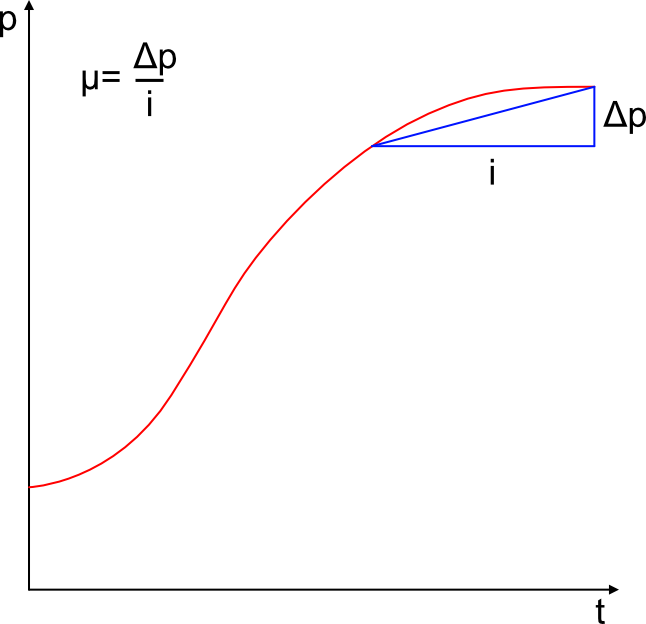}
   \end{minipage}
   \hspace{0.17\textwidth}
   \begin{minipage}{0.25\textwidth}
     \centering
     \includegraphics[width=\textwidth]{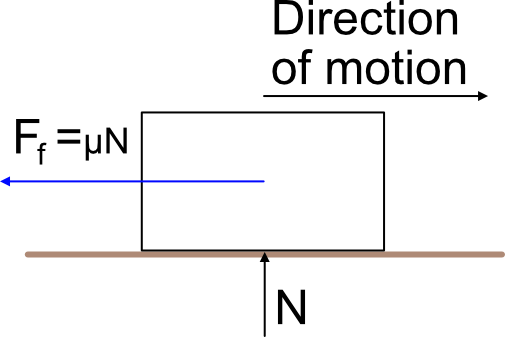}
   \end{minipage}
   
   \caption{Plot representing the \emph{Friction-based} progression. The change in performance of the agent is measured over an interval; this value is then used to calculate $F_f$, the friction force that slows down the object as the agent improves. This, in turn, results in the complexity of the environment increasing.} \label{fig:fbp_explanation}
\end{figure}

We introduce the Friction-based progression, which changes the complexity of the environment according to the agent's performance. This process, illustrated in Figure \ref{fig:fbp_explanation}, starts by measuring the change in the agent's performance in the last $i$ time-steps. This change is then divided by the magnitude of the interval to get the rate of change of the agent's performance over the interval, $\mu$. To utilise this information in creating a Curriculum, we introduce the model of a box sliding on a plane (Figure \ref{fig:fbp_explanation} right) and use its speed to calculate the environment's complexity. The intuition is that the relationship between speed and complexity is $complexity = 1 - speed$. This allows us to utilise $\mu$ as 
the friction coefficient between the body and the plane, resulting in the object slowing down and the complexity increasing as the agent's performance increases.


The intuition behind this progression function is that the more skilled the agent is at solving the current task, the quicker the agent should progress to a more difficult task. The physical model used provides a convenient way to translate a change in the agent's performance to a change in the difficulty of the environment, as can be seen in the equation below:






\begin{equation}
\label{eq:fbp}
\Pi_{fu}(t, m, s_{t-1}, s_{min}, p_{t-i}, {p_t}, i) = 1 - Uniform(s_t, s_{min})
\end{equation}
\begin{equation}
\label{eq:fbp.sp}
s_t = max(0, min(1, s_{t - 1} - m \cdot g \cdot \mu_t))
\end{equation}
\begin{equation}
\label{eq:fbp.fr}
\mu_t = \frac{p_t - p_{t-i}}{i}
\end{equation}

\noindent where $m$ is the mass of the object, $s_t$ is the object's speed at time $t$, $s_{min}$ is the minimum speed reached by the object, ${p_t}$ is the agent's performance at time $t$ and $i$ is the time interval considered by the progression function. Equations \ref{eq:fbp.sp} and \ref{eq:fbp.fr} describe our physical model whose speed is adjusted according to the agent's performance, while Equation \ref{eq:fbp} specifies the relationship between the speed of the model and the complexity of the environment. In order to respect the constraint that $c_t \in [0, 1]$, the speed of the object is clipped to fall within that same interval, as seen in Equation \ref{eq:fbp.sp}. The model used in our approach is not strictly physically accurate, as it considers the possibility of a ``negative friction" (the force is applied in the direction of the object's motion), where the speed of the object increases. This is the desired behaviour because if the progression function cannot decrease the complexity value, the agent might reach environments that are too difficult to solve. Without the possibility to revert to a solvable environment, the agent's policy and/or value function could deteriorate. In order to mitigate the risk of this happening, we introduced a uniform sampling in Equation \ref{eq:fbp}, which takes place whenever the agent's performance drops. A drop in performance would, in fact, result in the speed of the object increasing ($s_t > s_{min}$), which in turn would decrease the average complexity of the environment, but still allows the agent to train on a range of different complexities. The effect of the uniform sampling can be observed in Figure \ref{fig:prog}, especially when $i = 20$, where the complexity of the environment oscillates as a result of the performance decreasing. An example of the effect the value of $i$ has on the progression can also be seen in Figure \ref{fig:prog} (centre). In the same Figure, the plot on the right shows the performance function used to generate the central Figure. The performance function was chosen to have an underlying linearly increasing trend but masked by adding a periodic function. This performance function was used to showcase the behaviour of the Friction-based progression in a non-trivial scenario. This plot highlights how progressions with small intervals tend to react quickly to a change in the agent's performance. On the other hand, progressions with larger intervals provide a more constant increase in complexity.

Equation \ref{eq:fbp.fr}, used to calculate the friction coefficient $\mu_t$, relies on the performance function's past values, which implies that Equation \ref{eq:fbp} is only defined when $t = i + 1$. However, not changing the complexity value for the first $i$ time-steps would be highly detrimental to our approach, as the intervals used are often a not negligible percentage of the total training time. This is the motivation behind our choice to start the training at $t = i + 1$ and to consider the agent's performance for $t \leq i$ to be equivalent to the performance the agent would have when not performing any actions; this, in fact, allows us to start the progression immediately. A further motivation behind this choice lies in the fact that the only values of $p$ that influence when a progression ends ($c_t = 1$) are $\{p_1, ..., p_i\}$ and $\{p_{t-i}, ..., p_t\}$. 
This statement is supported by the following theorem, proven in Appendix A.

\noindent
{\bf Theorem} {\it Let $\bar{p}_{[x, y]}$ be the average value of the performance function between $t = x$ and $t = y$. Then the Friction-based progression with mass of the object $m$ ends when:

\begin{equation}
\label{eq:parSel}
\bar{p}_{[t+1-i, t]} - \bar{p}_{[1, i]} = \frac{1}{mg}
\end{equation}

} 
\noindent
The theorem above can also be used to simplify the parameter selection for our algorithm. Equation \ref{eq:parSel} can be solved for $m$ and be used to determine its appropriate value, resulting in the progression ending when a certain average performance is reached for $i$ time-steps. This results in the Friction-based progression only having two parameters in practice: the magnitude of the interval and the performance after which the agent should start training on the target task. In fact, $s_{t-1}$ and $s_{min}$ are initialised at one and kept in memory throughout the training, and $p_t$ is provided to the progression function at every time-step and stored in memory for $i$ iterations.

If the environment or the Curriculum were to require the progression to be monotonic, such as if some forms of transfer are involved, Equation \ref{eq:fbp} can be modified in the following way:

\begin{equation}
\label{eq:fbp_monotonic}
\Pi_{fm}(t, m, s_{t-1}, s_{min}, p_{t-i}, {p_t}, i) = 1 - s_{min}
\end{equation}

An example of a transfer learning algorithm that might require this monotonic progression is policy transfer using reward shaping \citep{brys2015policy}. This method, in fact, trains a randomly initialised policy by using a potential based shaping function influenced by the policy to be transferred. If this method of transfer was necessary once a complexity threshold was reached, and the agent's performance decreased, the uniform sampling in Equation \ref{eq:fbp} could result in the complexity decreasing below this threshold. This, in turn, would result in having to train a new randomly initialised agent, making it inefficient to reset the agent's weights potentially every episode. As mentioned above, Equation \ref{eq:fbp_monotonic} mitigates this problem by not allowing the complexity of the environment to decrease.

The space complexity of the Friction-based progression is $\Theta(i)$, as every value of the performance function in the interval needs to be stored. Moreover, the time complexity of one iteration of the Friction-based progression is $\Theta(1)$, as the value of $s_{min}$ is stored after every iteration and does not need to be re-computed. To accurately analyse our approach's time complexity, one also needs to consider the complexity of the mapping function, which is domain dependant. 


\subsubsection{Progression with parallel environments} \label{sec:parallel}

If it is possible to train with multiple instances of the domain simultaneously, like whenever using a parallel implementation of an RL algorithm, each instance of the domain should have its own independent progression function. Having multiple progression functions allows for varying their parameters, resulting in the agent training on a range of complexities of the environment simultaneously. This also varies the experience of different learners and results in a higher stability of the training process, as the correlation of the updates between different learners is reduced. This technique applies to both the Exponential and Friction-based progression functions.

When using the Exponential progression, all learners use the same $t_e$ but vary the value of $s$, like in Figure \ref{fig:prog}. Once the number of processes is specified, a minimum and a maximum value for $s$ are defined; firstly, two processes are assigned $s$ corresponding to the minimum and the maximum. The rest of the processes then are assigned a value of $s$ that ensures even spacing from the line $x = y$. For example, the four lines with $s > 0$ in Figure \ref{fig:prog} would result from this parameter selection method using 4 processes. This technique also has the effect of requiring less variance in the parameter selection process. In fact, in all of the domains where this technique is applicable, we used the same parameters for selecting $s$: the minimum value of $s$ was always 0.1, and the maximum value of $s$ was 2.

The benefits of this technique lie in the complementary nature of progression functions with long and short intervals when using the Friction-based Progression. As mentioned previously, progressions with a small interval tend to react more suddenly to a change in the agent's performance; progressions with a higher interval, on the other hand, provide a smoother increase in complexity. Having multiple progression functions simultaneously allows us to use a wide spectrum of intervals simultaneously. An example of this can be seen in Figure \ref{fig:prog} where 4 different progressions are plotted using the same performance function. In practice, the performance would not be the same for all progression functions, as different learners would also be training in environments of different complexities.

\subsection{Mapping Functions}

Alongside Progression functions, our framework also relies upon Mapping functions in order to generate a Curriculum. As previously mentioned, Mapping functions cover the task generation aspect of a Curriculum. In fact, once the progression function specifies the desired complexity, the role of the mapping function in our framework is to generate the MDP associated with that complexity. This can be seen in Algorithm \ref{algo:clpf} at line 12, where $M_i$ is the $i^{th}$ MDP in the Curriculum.

Mapping functions are also how high-level domain knowledge can be leveraged in creating a Curriculum: to demonstrate this, we expand on the example previously introduced in Section \ref{sec:map_and_prog}, which dealt with creating a mapping function for navigation domain. Let us assume the domain in question requires one parameter to define an MDP: the distance at which the agent should be initialised from the goal. Let us also assume that this distance can take any value between 0 and 10. Finally, let the domain knowledge we want the Mapping function to represent be: the further away from the goal the agent is initialised, the more complex the MDP. This knowledge can be encoded by this simple mapping function:

\begin{equation}
\label{eq:mapping_example}
\Phi(c_t) = M_{c_t * 10}
\end{equation}

\noindent where $M_x$ is the MDP where the agent's initial distance from the goal is $x$. In this example, there is a linear correlation between complexity and distance, although, as shown in the results section, when using the Friction-based progression, as long as the proposed mapping function does not violate the domain knowledge, the specific way the mapping function is defined does not greatly affect the performance. In this specific domain, violating domain knowledge would imply considering MDPs where the agent starts closer to the goal as being more complex than MDPs where the agent starts farther from the goal. As the relationship between the complexity and the value of the parameters in the mapping function has a limited effect on the agent's performance when using the Friction-based progression, to keep Mapping functions as simple as possible, we encourage a linear correlation between the two.

If the monotonicity of a parameter's complexity can be assumed, the value of this parameter relative to the environment's complexity can be easily calculated using one of the two equations below:



\begin{equation}
\label{eq:maze_mapping_h}
a_h > a_e \rightarrow a_t = a_e + (a_h - a_e) * c_t
\end{equation}
\begin{equation}
\label{eq:maze_mapping_e}
a_e > a_h \rightarrow a_t = a_e - (a_e - a_h) * c_t
\end{equation}

\noindent
where $a_e$ is the value of parameter $a$ corresponding to the most complex environment, $a_h$ is the value of parameter $a$ corresponding to the least complex environment, $a_t$ is the value of parameter $a$ at time $t$, and $c_t$ is the complexity value at time $t$. These two simple equations were used to generate all of the mapping functions in our domains and are applied to each parameter independently. The advantage of generating mapping functions using the equations above is that if the assumption holds, the only thing that needs to be ascertained is whether the minimum value of a parameter $a_{min}$ results in an easier or harder environment compared to the maximum value of the same parameter $a_{max}$. If the assumption did not hold naturally for a domain, for example, if the domain used as an example earlier required to specify the $x$ and $y$ coordinates, we would introduce distance as an abstraction of the coordinates of a state, satisfying the constraint above. If a domain did not allow for such an abstraction, the mapping function could still be generated not to violate the domain knowledge, although without using Equations \ref{eq:maze_mapping_h} and \ref{eq:maze_mapping_e}.

\section{Experimental Evaluation}


As mentioned in the introduction, our method sits at the crossroads between Curriculum Learning algorithms that focus on sequencing and modifying the agent's experience within a given task. Our goal, therefore, is to compare our method against one state-of-the-art algorithm from each class: HTS-CR and Reverse Curriculum Generation. For the justification of the choice of each method in its corresponding class refer to Section \ref{sec:related_work}. This section describes the details of our evaluation, first introducing the six domains used for our comparison, then providing some details on the implementation of our algorithm and the two baselines selected to aid reproducibility.

\subsection{Evaluation domains}

\begin{figure*}[!t]
\centering
   \begin{minipage}{0.32\textwidth}
   \centering
     \includegraphics[width=\textwidth, height=\textwidth]{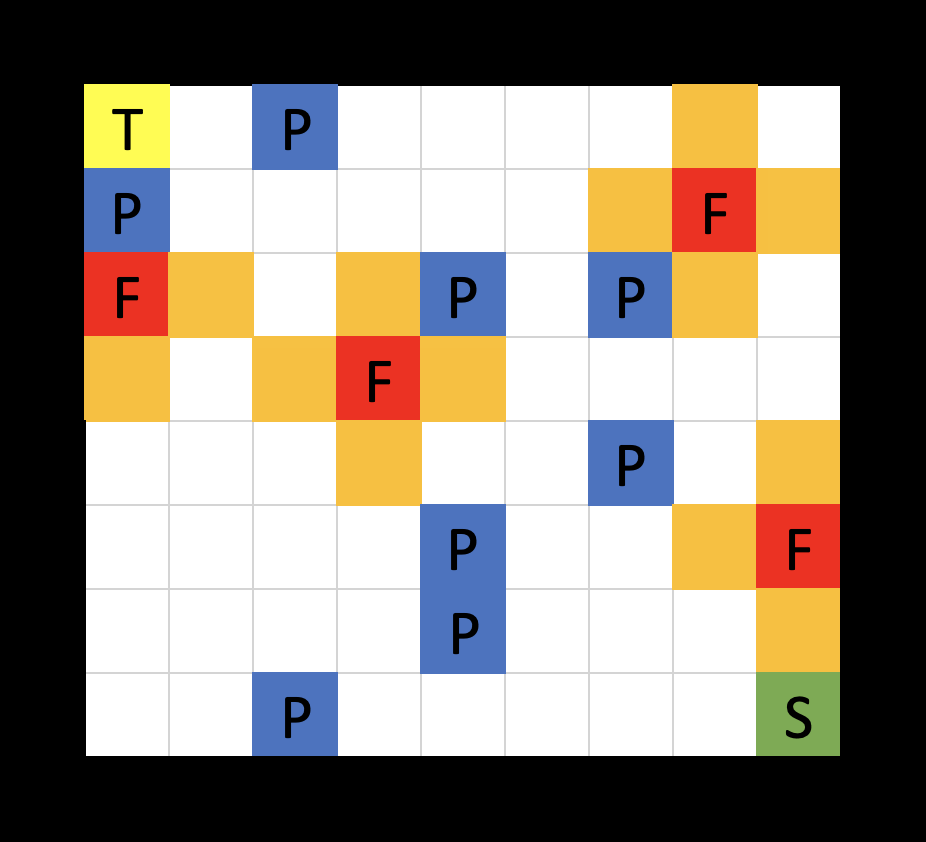}
   \end{minipage}
   \begin{minipage}{0.32\textwidth}
     \centering
     \includegraphics[width=\textwidth, height=\textwidth]{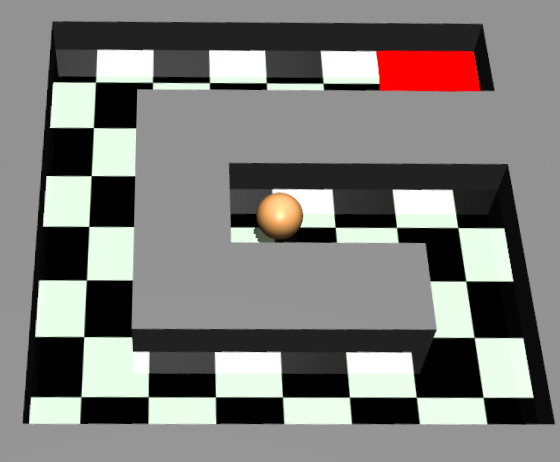}
   \end{minipage}
   \begin{minipage}{0.32\textwidth}
     \centering
     \includegraphics[width=\textwidth, height=\textwidth]{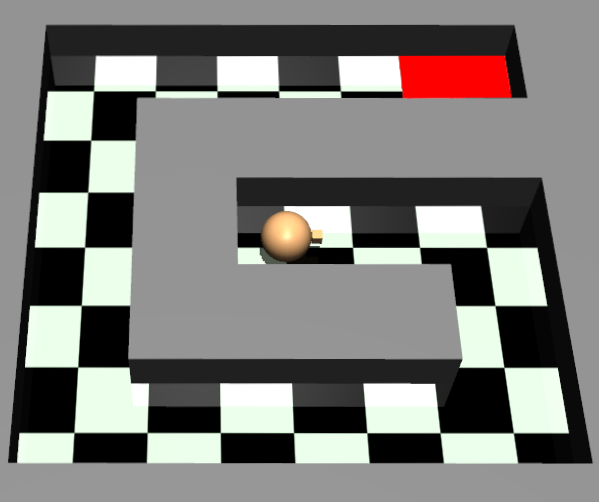}
   \end{minipage}
   
   \vspace{0.005\textwidth}
   
   \begin{minipage}{0.32\textwidth}
     \centering
     \includegraphics[width=\textwidth, height=\textwidth]{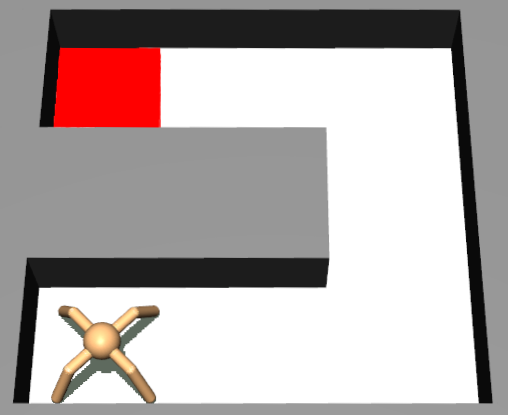}
   \end{minipage}
   \begin{minipage}{0.32\textwidth}
     \centering
     \includegraphics[width=\textwidth, height=\textwidth]{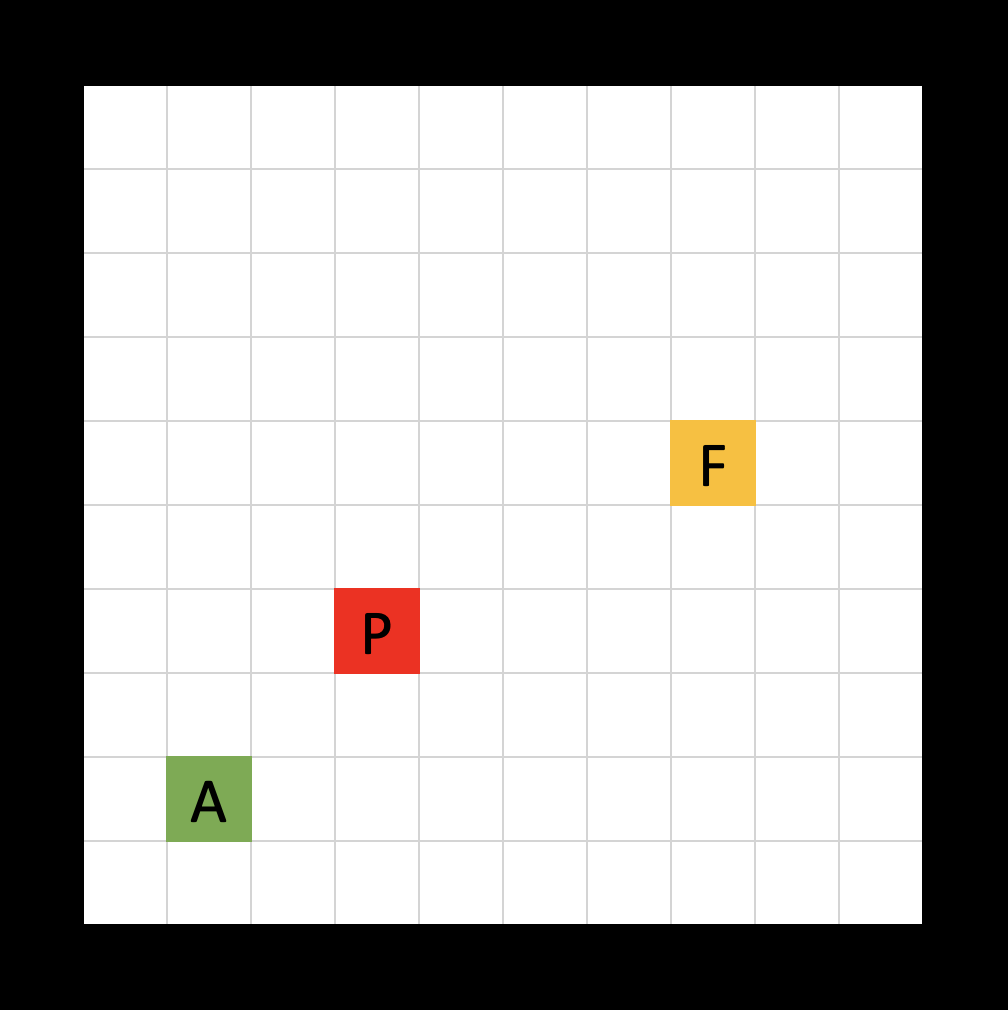}
   \end{minipage}
   \begin{minipage}{0.32\textwidth}
     \centering
     \includegraphics[width=\textwidth, height=\textwidth]{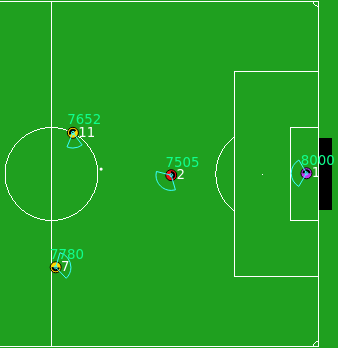}
   \end{minipage}
   
   \caption[asdf]{Visualisation of each one of the domains used in our testing. (From top left to bottom right: Grid World Maze, Point Mass Maze, Directional Point Maze, Ant Maze, Predator-Prey and HFO\footnotemark)} \label{fig:visual}
\end{figure*}

Our testing domains include two Grid World environments, three MuJoCo environments and Half Field Offense. A visualisation for each domain is available in Figure \ref{fig:visual}.

\footnotetext{Figure for the HFO environment edited from \url{https://github.com/LARG/HFO/blob/master/img/hfo3on3.png}}

\paragraph{Grid World Maze}

The first test domain used in this paper is a grid-world domain previously used by \cite{foglino2019curriculum}, where the agent needs to reach a treasure while avoiding fires and pits. The agent can navigate this domain by moving North, East, South or West, and the actions taken in this domain are deterministic. This domain's reward function is as follows: 200 for reaching the cell with the treasure, -2500 for entering a pit, -500 for entering a fire, -250 for entering one of the four adjacent cells to a fire and -1 otherwise. In this domain, the episodes terminate when the agent reaches the treasure, falls into a pit or performs 50 actions. The combination of fires and pits used can be seen in Figure \ref{fig:visual}, where \textbf{F} represents a fire, orange squares represent a cell next to a fire, \textbf{P} represents a pit, \textbf{S} represents the starting position for the target task, and \textbf{T} represents the position of the treasure.


\paragraph{MuJoCo Mazes}

Three additional domains are MuJoCo \citep{todorov2012mujoco} environments where the agent needs to navigate a maze to reach a goal area. The environment's complexity is determined by two factors: the model the agent controls and the maze the agent needs to solve. We use two configurations from \cite{florensa2017} and introduce a new configuration as a further domain.

The first model, Point Mass, is a sphere controlled by modifying its acceleration on the x and y-axis. The second model, Directional Point, is a sphere with a cube added to one end of the point as a directional marker. The point has two degrees of freedom: it accelerates in the marker's direction and rotates around its own axis.
Finally, the third model, Ant, is a sphere with four limbs, with a joint each, attached to it. This results in the model having 8 degrees of freedom and the navigation task being very challenging.

We used two different mazes in our testing process, the more simple one shaped like an inverted C, with the goal in the top left corner, the second one being a G shaped maze with the goal in the top right corner. As previously mentioned, we used three distinct configurations of maze shape and agent models: like in \cite{florensa2017}, the Point Mass model was paired with the G shaped maze, and the Ant model was paired with the C shaped maze. We also introduced a new combination, the Directional Point model paired with the G shaped maze, to add a navigation task of intermediate complexity to our testing environments.

In these environments, the agent receives a reward of 1 when entering the goal area and 0 otherwise. The episodes terminate when the agent reaches the goal or after 300 (Point Mass and Directional Point) or 2000 (Ant) time-steps.

A visualisation of these environments can be seen in Figure \ref{fig:visual}, where the agent is shown in the initial position associated with the target task, and the area highlighted in red is the goal the agent learns to reach.

\paragraph{Predator Prey}

A further domain is a grid-world domain where the agent needs to survive by eating stationary food sources while escaping a predator. The agent starts with 100 health points and loses one health point per time-step and a further 100 health points if a predator eats the agent. Eating food restores 10 health points and spawns a new food source at a random location, and the agent reaching 0 health points terminates the episode. The reward function used is the difference between the agent's health before and after each action. The state is represented as an 11 by 11 three-channel array, and the agent can move in one of the four cardinal directions. An agent taking an action that would result in an out of bounds position results in the agent being stationary for that time step. If the agent can survive for 1000 time-steps, the episode terminates without further positive or negative reward. This domain can be observed in Figure \ref{fig:visual}, where \textbf{A} represents the agent, \textbf{P} represents the predator, and \textbf{F} represents a food source.

\paragraph{Half Field Offense}

The final testing domain is Half Field Offense, a subtask in RoboCup simulated soccer \citep{hausknecht2016half}. In this domain, the agent controls the closest player to the ball and belonging to the attacking team on a football pitch to score a goal against the defending team. The playing area is restricted to half of the pitch, and the number of players on each team is set to two.
The state-space for this environment is the "High level" state space described in \cite{hausknecht2016half}, which is composed of $24$ features. The action space includes five discrete actions: shoot, pass, dribble, move and go to the ball. A further layer of complexity is added because if the agent tries to pass, dribble or shoot without being near the ball, the agent will take no action for that time step.
The reward function awards the agent a utility of 1 for scoring a goal and -1 when the ball goes out of bounds, is captured by the defending team, or the episode lasts for longer than 500 time-steps. This domain can be seen in the bottom right corner of Figure \ref{fig:visual}.





\subsection{Experimental setting}

This section includes the details of the implementation of the various algorithms used in our evaluation.

\subsubsection{Mapping and Progression Functions} \label{sec:progression}

In all of the environments used for our testing, we always had a linear correlation between the complexity and value of the parameters in our evaluation domains. Moreover, in all of the environments used for testing, the complexity relative to each parameter was monotonic (note that this is not a requirement of our framework), making it possible to generate mapping functions by simply using equations \ref{eq:maze_mapping_h} and \ref{eq:maze_mapping_e}.

\paragraph{Navigation domains} In all of our navigation domains, the only parameter used to generate MDPs was the distance at which the agent was initialised from the goal. In the Grid World Maze domain, we also specified that this random state could not be associated with a negative reward (in a pit, on or next to a fire). Given a certain $c_t$, the distance to the goal was calculated using Equation \ref{eq:maze_mapping_h}, where $a_t$ would be the resulting distance.

\paragraph{Predator prey}In the Predator-Prey domain, two parameters need to be specified: the percentage of squares occupied by prey and the number of time-steps the predator stays still after attacking the agent. The first parameter was included in our mapping function using Equation \ref{eq:maze_mapping_e}, whereas the second parameter was calculated using Equation \ref{eq:maze_mapping_h}.

\paragraph{HFO} In the Half Field Offense domain, the MDPs created by the mapping function modified three parameters: the proximity of the initial position of the ball to the halfway line, how close the ball could be initialised to the sideline, and whether the attacking team started with the ball in their possession. The mapping function was generated using Equation \ref{eq:maze_mapping_e} for the first parameter and Equation \ref{eq:maze_mapping_h} for the second parameter. On the other hand, the third parameter has a peculiar property: it can only assume two values. When such a parameter is found, the equations defined above will prescribe to switch its value once the complexity reaches a value of 0.5. This might pose a problem when dealing with parameters that can only assume two values. Changing their value might, in fact, significantly increase the complexity of the environment, and being able to control where this happens during the training is necessary. Therefore, in our mapping function, we let the attacking team start with the ball until $c_t$ surpasses 0.2. This allows the agent to initially gather some positive reward from successfully scoring goals and subsequently learn to go to the ball before taking any other actions. This sudden change in the environment is better suited to the initial part of the training when the agent tends to explore more, and its learning rate tends to be higher.

In all of the domains mentioned above, except for Half Field Offense, the agent was trained using multiprocessing. As mentioned in Section \ref{sec:parallel}, this allowed us to use multiple instances of progression functions with different parameters to vary the experience gathered by different parallel workers. When using this technique on the Exponential Progression, the set of parameters used for all of the environments that could be parallelised was the same. The time at which the progression should end, $t_e$, was always set at $80\%$ of the total training time, and the minimum and maximum values of $s$ were respectively 0.2 and 2. Successfully utilising the same set of parameters on multiple environments highlights the robustness of our approach. On the other hand, when applying the Exponential Progression to Half Field Offense, where the implementation did not include multiprocessing, the value of $t_e$ chosen was still equivalent to $80\%$ of the total training time. In contrast, the value of $s$ used was 1.

When using this technique with the Friction-based Progression, the
the aim is to vary the magnitude of the interval as much as possible between different workers. To keep our parameter selection consistent between domains, wherever we trained with more than four concurrent workers, the maximum value of the interval was 10 times the minimum value of the interval. On the other hand, in the Grid World Maze domain, as we trained using 4 concurrent workers, we set the maximum interval to be 3 times the minimum interval. 

To utilise the Friction-based progression, it is necessary to define a performance function that varies from domain to domain. All the MuJoCo navigation domains used the Cumulative Return as their performance function, whereas in the Grid World Maze domain, the Cumulative return was chosen but clipped to zero. That is, if the cumulative reward for the episode turned out to be negative, the progression function would consider its value as 0. This technique is useful to stabilise the progression in the Grid World Maze environment, where the negative reward for falling into a pit is one order of magnitude bigger than the reward for reaching the goal. On the Predator-Prey domain, the performance function chosen was the episode duration, as the agent's survival is the main objective in this environment. Finally, in the HFO domain, the performance function returned 1 if the agent scored a goal and 0 otherwise.

The last variable related to the Friction-based progression is the performance needed on the final task in order for the progression to end. In the MuJoCo navigation domains and HFO, this was set at half the performance achievable using an optimal policy. On the other hand, this was set at the minimum cumulative return achievable by a successful episode on the Grid World Maze domain. Finally, in the Predator-Prey domain, this value was set at a value between the optimal policy and how long an agent would survive without gathering any food but just escaping the predator.

Finally, as the linear progression can be approximated using an exponential progression as shown in Equation \ref{eq:linexpeq}, our experimental section will only include the latter.

\subsubsection{Baselines} \label{sec:rcg_impl}

\paragraph{Reverse Curriculum Generation}The version of the Reverse Curriculum Generation used as a baseline for our approach generates a Curriculum via a random sampling of actions from starting states solvable by the agent but that have not been yet mastered. This algorithm also includes a replay buffer that considers the agent's performance over the course of the training to propose starting states to be replayed.

As this algorithm generates a Curriculum by modifying the starting state and assumes that given two states $a$ and $b$ in a certain MDP, you can get from $a$ to $b$ using a uniform sampling of random actions, it could not be applied to two of our testing domains: HFO and Predator-Prey. They both contain an AI that plays against the agent with a specific policy that defends the goal (HFO) or chases the agent (Predator-Prey). This violates the assumption mentioned above: in HFO, let state $a$ be a normal starting state for the domain, where the agents are initialised next to the halfway line, and the defence team is initialised next to the goal. Let state $b$ be the state where the agents are next to the goal, and the defenders are next to the halfway line. The agents can take no sequence of actions that will result in state $b$ occurring when the environment is initialised in state $a$, as whilst the agents might get close to the goal, the defence team would never abandon their task and go next to the halfway line. Likewise, in the Predator-prey domain, a similar issue arises where the predator would keep following the agent, limiting the portion of the state space that can be reached from a certain starting state.

As RCG creates a Curriculum by only modifying the agent's starting state, in the domains where RCG is applicable, our algorithm's mapping function only modifies the agent's starting state.

\paragraph{HTS-CR} Using HTS-CR \citep{foglino2019curriculum} as one of our baselines, we used transfer between tasks; the transfer method consisted of directly transferring the neural network from one task to the next, using it as a better initialisation on the next task. One limitation of HTS-CR is that it cannot use the full set of intermediate tasks used in our algorithm, like any other task sequencing algorithm. In fact, this set is often infinite and, if used in its entirety, would result in an evaluation only containing Curricula of length 2. To provide a fair comparison, the set of intermediate tasks was selected from a subset of the tasks used by our approach. More specifically, the set of source tasks considered by HTS-CR is $<\Phi_{D}(0), \Phi_{D}(0.25), \Phi_{D}(0.5), \Phi_{D}(0.75)>$, with the target task being $\Phi_{D}(1)$. Four tasks were included to ensure HTS-CR had enough time to complete the head and tail selection process and evaluate some longer curricula within the time limits. The comparison against this baseline helps to ascertain whether our approach using a continuously changing intermediate task can outperform what is achievable with a limited set of tasks. HTS-CR was modified to optimise the performance on the final task rather than the cumulative return; this does not violate the guarantee of optimality and is the more appropriate metric to be optimised for this comparison. HTS-CR was trained 10 times the amount compared to the other algorithms in all of our test domains. In the plots provided in the results section, the final performance of HTS-CR is reported as a dashed green line.
As at every iteration, HTS-CR starts training on the final task only after having trained on the full curriculum; when plotting the performance of this algorithm, its curve will not start until later in the training.

\subsubsection{Learning algorithm} All the domains in this paper use the same learning algorithm: Proximal Policy Optimisation \citep{schulman2017proximal}, in particular, the implementation contained in the ``Stable-baselines" package \citep{stable-baselines}. Moreover in all the domains except for HFO, the parallel implementation of the method was utilised, where the experience collected by multiple independent workers is used to update the policy.

\subsubsection{Experimental method} To provide statistically reliable results, each algorithm except for HTS-CR was executed 50 times for each domain except for Predator-Prey, where each algorithm was executed 20 times. As each evaluation made by HTS-CR was the average of 5 runs, the number of runs for this algorithm on each domain was divided by 5. The results section will then report the mean and $95\%$ confidence interval for each algorithm.

The metric of comparison chosen for the Grid World domain was the cumulative return from the designated starting coordinates. In contrast, the metric of comparison chosen for the MuJoCo navigation domains was the probability of success from a uniform sampling of the states in the maze, previously used by \cite{florensa2017}. In the Predator-prey domain, the survival time was chosen as the evaluation metric; finally, in HFO, the metric used was the probability of scoring a goal.

\section{Results}

This section presents the results on the six testing domains. First we perform an ablation study to establish the role of several components of the framework: uniform sampling of the friction-based progression, parallelism among learners, and correctness of the mapping function. We aim to verify experimentally that uniform sampling is beneficial to the agent's performance; that parallel learning does not result in a (possibly faster) convergence to a worse behaviour than with single-process learning; and that the friction-based progression is robust to errors in the mapping function.
Finally, with the last experiment we evaluate our framework by comparing the Exponential and Friction-based progression to other state-of-the-art Curriculum Learning algorithms.

\subsection{Friction-based progression}

This section seeks to verify that uniform sampling in the friction-based progression has a positive effect on the agent's performance.

In order to verify whether the uniform sampling introduced in Equation \ref{eq:fbp} is advantageous, we compared this formulation of the Friction-based progression, referred to as \emph{uniform} from here onward, to the \emph{monotonic} formulation introduced in Equation \ref{eq:fbp_monotonic}. We also developed an alternative formulation aimed at modelling the more simple case where there is a direct correlation between the speed of the object and the complexity of the environment, modelled by the following equation:

\begin{equation}
\label{eq:fbp_speed}
\Pi_{fu}(t, m, s_{t-1}, p_{t-i}, {p_t}, i) = 1 - s_t
\end{equation}

\noindent Given the direct relationship between the speed of the object and performance of the agent, this formulation will be referred to as \emph{speed} from here onward.

\paragraph{Experimental results}

We compared the three formulations described above on all our six testing domains to find the best performing one and explore each formulation's properties. The plots for these experiments can be found in Figure \ref{fig:noise_fbp_variants}, with Table \ref{tab:variants} reporting the highest performance achieved on each domain by each formulation. 

Our experimental results show that in the Point Mass Maze environment, there is a clear benefit when using the \emph{speed} formulation, as it results in a significant increase in performance during the initial part of the training and also results in a much quicker convergence to the optimum. In this domain, the \emph{sampling} and \emph{monotonic} formulations perform similarly, with the \emph{sampling} formulation resulting in slightly higher performance at the start of the training and converging within the confidence interval of the best performing formulation.

In both the Directional Point Maze domain and the Ant Maze domain, all three formulations perform equally both throughout the training and as far as the best performance achieved is concerned.

When analysing the results in the Grid World Maze, HFO and Predator-prey domains, the benefits of the \emph{sampling} formulation are highlighted over the \emph{speed} formulation. Firstly in the Grid World Maze domain, the \emph{sampling} and \emph{monotonic} formulations perform equally throughout the training, whereas the \emph{speed} formulation has a lower probability of solving the maze in the initial part of the training. Moreover, the \emph{speed} formulation converges to a considerably lower average value, although still within a $95\%$ confidence interval to the other two formulations. Moreover, in the HFO domain, the \emph{speed} formulation results in a lower initial performance compared to the other two. This is significant as the discrete parameter included in this domain, whether the offending team starts with the ball, is changed towards the start of the progression. In the Predator-prey domain, on the other hand, the performance of the \emph{speed} formulation drops significantly in the later stages of the training, unlike the other two formulations. This is due to instability in the agent's performance as the progression gets to the target task, resulting in the progression quickly lowering the complexity of the environment. Not being able to train on the hardest complexity directly, in turn, hurts the testing performance of the agent on the target task, which is particularly evident when looking at Figure \ref{fig:noise_fbp_variants}. This is mitigated by the \emph{monotonic} formulation by having the agent train on the target task directly whenever the progression finishes; however, the experimental results suggest that the \emph{sampling} technique is the most effective. This pattern is also present in the HFO domain, where the \emph{sampling} formulation performs significantly better than the \emph{monotonic} progression.

Based on our experiments, we hypothesise that the definition of the Friction-based progression in Equation \ref{eq:fbp_speed} results in the algorithm having a weakness in domains with discrete parameters that can significantly increase its difficulty. In fact, if the agent's performance drops significantly in the harder environment, the progression would immediately decrease the complexity, and the agent would not have enough time to learn this new and more complex environment. This could, in turn, result in the progression being halted. This is supported by the fact that whenever an environment has a discrete parameter, the performance of the \emph{speed} formulation was inferior compared to when using a uniform sampling.

Overall our experimental results highlight the benefits of the higher stability provided by the \emph{sampling} formulation, which was always either the best performing one or within the confidence interval of the best performing formulation. The \emph{speed} formulation seems to be best suited for less complex domains and seems to be less effective whenever there are discrete parameters in the domain. Finally, the \emph{monotonic} formulation seems to be more stable than the \emph{speed} formulation and would be useful if a monotonic increase in complexity was needed. As the \emph{sampling} formulation was found to be the best performing one, from this point onward, whenever we refer to the Friction-based progression, the \emph{sampling} formulation is implied.



\begin{table}[H]
 \small
 \centering
 \begin{tabular}[t]{ccccccc} 
 \hline
 Variant & GW Maze & Point Maze & Dir. P. M. & Ant Maze & 2v2 HFO & Pred. Prey\\ [0.5ex] 
 \hline
 Speed & \bf89.4 $\pm$ \bf6.0 & \bf100 $\pm$ \bf0.0 & \bf98.5 $\pm$ \bf0.6 & \bf95.0 $\pm$ \bf1.5 & 65.7 $\pm$ 6.1 & 216.8 $\pm$ 13.3\\
 
 Sampling & \bf94.4 $\pm$ \bf3.9 & \bf99.7 $\pm$ \bf0.4 & \bf97.8 $\pm$ \bf1.2 & \bf95.1 $\pm$ \bf1.4 & \bf75.8 $\pm$ \bf1.4 & \bf249.8 $\pm$ \bf8.5\\
 
 Monotonic & \bf94.0 $\pm$ \bf3.6 & 99.1 $\pm$ 0.7 & \bf98.2 $\pm$ \bf0.7 & \bf94.6 $\pm$ \bf1.0 & 68.0 $\pm$ 5.0 & \bf238.2 $\pm$ \bf15.6\\
 
 \hline
\end{tabular}\\
\caption{Best performance on each domain. All the methods within a $95\%$ confidence interval from the best performing method are in bold.} \label{tab:variants}
\end{table}

\subsection{Effects of Multiprocessing}

In this experiment, we seek to verify whether the use of multiprocessing hinders our approach's performance. Further to this, we also want to verify if multiprocessing has any added benefits aside from speeding up the learning process.

In order to verify the hypothesis stated above, we tested the Exponential and Friction-based progression on our testing domains. Both progression functions were tested with and without multiprocessing in order to be able to compare their performance and draw some more general conclusions on the effects of multiprocessing on our method. Because of the instability of the code used in the HFO domain, it was not practical to use multiprocessing in this case; thus, we will only be performing this test on the remaining five domains.

\paragraph{Experimental results}


Our experimental results show a clear increase in performance over the first half of the training when using multiprocessing for both progression functions. This increase in performance is sustained throughout the training for the Friction-based progression; however, the best average performance achieved by the exponential progression is achieved without multiprocessing. It is worth mentioning that this performance's confidence interval intersects the confidence interval relative to the method with multiprocessing.


In the Point Maze and Directional Point Maze domains, using multiprocessing in conjunction with the Friction-based progression results in increased average performance, although within a confidence interval of the Friction-based progression without multiprocessing. On the other hand, these domains resulted in a significant increase in the performance of the Exponential progression throughout the training.

There is no measurable difference between the Friction-based progression with or without multiprocessing on the Ant maze domain. On the other hand, using multiprocessing with the Exponential progression results in an increase in performance over the first half of the training but results in convergence to a similar value compared to the method without multiprocessing.

Finally, both methods converge to a higher average value using multiprocessing in the Predator-prey domain but converge just within the confidence interval of the same method without multiprocessing. The Friction-based progression without multiprocessing results in a higher average performance in the first half of the training but has a lower average performance for the rest of the training. On the other hand, the exponential progression without multiprocessing results in a lower performance from the first third of the training onward, although usually within the confidence interval of the Exponential progression with multiprocessing. One final detail that should be mentioned is that both the Friction-based progression and the Exponential progression without multiprocessing experienced a significant drop in performance at some point in the training, which is not present in their counterpart when using multiprocessing. This coincides with the time where the progression is set to end with the exponential progression and is likely also linked to a transition to the target task in the friction-based progression. This is a crucial time of the progression, and in this specific domain, this drop highlights that more stability is needed. In the case of multiprocessing, this stability is provided by having different processes end the progression at different times (Friction-based) or progressing to more complex environments with different schedules.

Our results suggest that as well as providing an overall speedup of the training, utilising multiprocessing as part of the Curriculum can result in an increase in performance. Concerning the Friction-based progression, we did not observe an instance in our testing where it was detrimental to use multiprocessing. On the other hand, when using an Exponential progression, the version without multiprocessing resulted in a higher best performance in the grid world maze domain, albeit still within the confidence interval of the version with Multiprocessing. This is offset by the Exponential progression benefiting from the use of Multiprocessing in our other four testing domains.

\begin{table}[H]
 \small
 \centering
 \begin{tabular}[t]{cccccc} 
 \hline
 Variant & GW Maze & Point Maze & Dir. P. M. & Ant Maze & Pred. Prey\\ [0.5ex] 
 \hline
 F. B. P. multi & \bf94.4 $\pm$ \bf3.9 & \bf99.7 $\pm$ \bf0.4 & \bf97.8 $\pm$ \bf1.2 & \bf95.0 $\pm$ \bf1.4 & \bf249.8 $\pm$ \bf8.5\\
 
 F. B. P. single & 74.6 $\pm$ 11.5 & \bf99.4 $\pm$  \bf0.5 & \bf97.6 $\pm$ \bf2.0 & \bf95.0 $\pm$ \bf1.4  & \bf239.6 $\pm$ \bf10.2\\
 
 \hline
 \hline
 
 
 Exponential multi & \bf77.5 $\pm$ \bf10.1 & \bf97.5 $\pm$ \bf1.7 & \bf79.3 $\pm$ \bf2.9 & \bf92.0 $\pm$ \bf2.0 & \bf239.3 $\pm$ \bf7.9\\
 
 Exponential single & \bf88.9 $\pm$ \bf7.8 & 75.3 $\pm$ 1.0 & 51.6 $\pm$ 1.5 & \bf91.0 $\pm$ \bf2.5  & \bf224.2 $\pm$ \bf8.3\\
 
 \hline
\end{tabular}
\caption{Best performance on each domain with and without multiprocessing, divided by method. If the algorithm with and without multiprocessing perform within a 95\% confidence interval, both performances are in bold, otherwise the highest of the two is in bold.}
\end{table}

\subsection{Robustness of the Friction-Based progression}

As progression functions rely on mapping functions in order to generate a Curriculum, we consider whether the Friction-Based progression is resilient to a badly formulated mapping function. This section is specific to the Friction-based progression as it is generated online and therefore can adapt to noise in the progression. Since the Exponential progression is determined before execution, in the case of a badly formulated mapping, its parameters would need to be modified on a case by case basis.

\begin{figure*}[!t]
\centering
   \begin{minipage}{0.32\textwidth}
     \includegraphics[width=\textwidth]{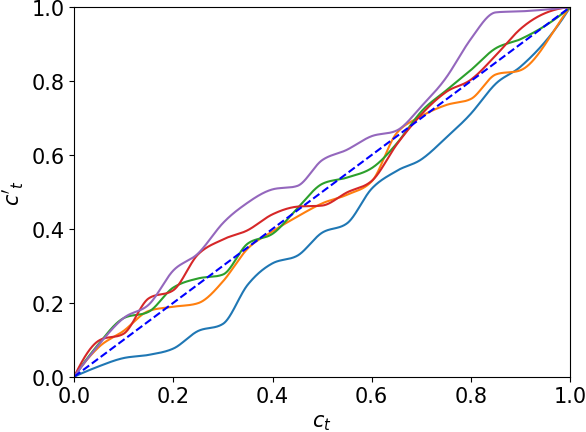}
   \end{minipage}
   \begin{minipage}{0.32\textwidth}
     \centering
     \includegraphics[width=\textwidth]{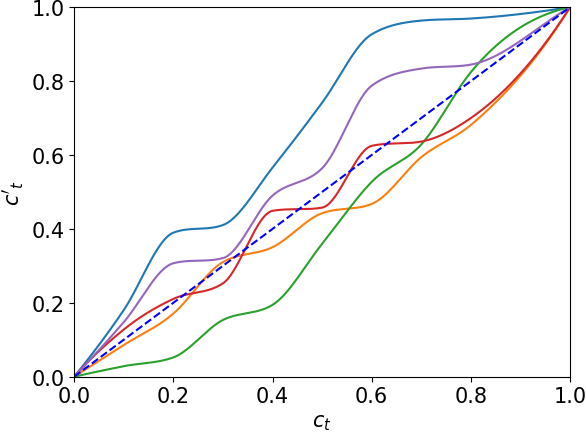}
   \end{minipage}
   \begin{minipage}{0.32\textwidth}
     \centering
     \includegraphics[width=\textwidth]{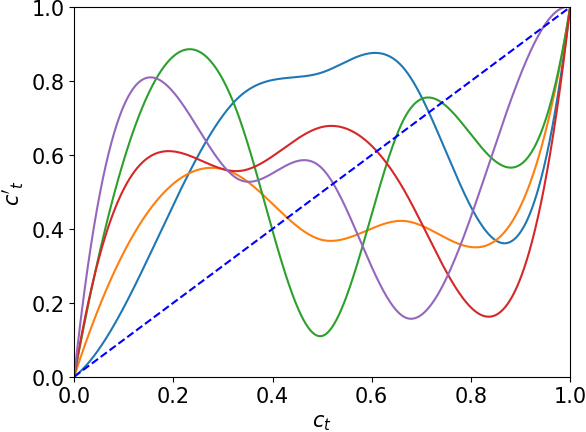}
   \end{minipage}
   \caption{Examples of the three different types of noise: Local (left), Global (center) and Random (right). The blue dashed line represents $c'_t = c_t$, i.e. no noise.}
   \label{fig:noise}
\end{figure*}

In order to assess the robustness of our method, we modelled a badly formulated mapping function as noise in the creation of a Curriculum. This noise takes the form of a function $\mathcal{N} : [0, 1] \rightarrow [0, 1]$ that maps the complexity factor $c_t$ to its noisy counterpart $c'_t$. This value will then be used in the generation of the MDPs to be added to the Curriculum, resulting in the noisy curriculum $C'$ being defined by the following two equations:

\begin{equation}
\label{eq:noisy_mdp}
M'_t = \Phi(\mathcal{N}(c_t))
\end{equation}
\begin{equation}
 \label{eq:noisy_curr}
C' = \langle M'_0, ..., M'_i\rangle 
\end{equation}


We identified two possible cases of mistakenly creating a mapping function. The first case is when the mapping function correctly identifies the domain knowledge that should be represented but encodes it incorrectly. For example, in a navigation domain, one such mapping function would correctly assume that a state closer to the goal would be at least as easy to solve as a state further away from the goal; however, the difference in complexity of these two states would be random. On the other hand, the second case is a random mapping function, modelling whenever even the domain knowledge encoded in the Mapping function is incorrect. In a navigation domain, this would result in a state closer to the goal being potentially considered more complex than a state further from the goal.

The first case results in the noise function being monotonic, implying that given two values $a$ and $b$, if $a > b$ then $\mathcal{N}(a) \geq \mathcal{N}(b)$. In this evaluation we defined two possible noise functions belonging to this class: \emph{local} and \emph{global}. The first type of noise is useful to assess how robust the Friction Based progression is to a sharp but localised perturbation, whereas the second type of noise tests the algorithm's ability to react to smoother but greater noise. These two types of noise can be seen in Figure \ref{fig:noise}. These two noise functions still need to respect the constraints below.

\begin{equation}
\label{eq:robustness}
a > b \rightarrow \mathcal{N}(a) \geq \mathcal{N}(b) \quad \forall \ a, b \in [0, 1]
\end{equation}
\begin{equation}
\label{eq:robustness1}
\mathcal{N}(0) = 0
\end{equation}
\begin{equation}
\label{eq:robustness2}
\mathcal{N}(1) = 1
\end{equation}

\noindent with the constraints in eq. \ref{eq:robustness1} and \ref{eq:robustness2} being necessary to ensure that all complexity values are still reachable. The algorithm used to generate $\mathcal{N}$ starts with the creation of a list $L = \{l_0, ..., l_n\}$ of $n$ random values sampled from a truncated normal distribution bounded between 0 and 1 with a mean of 0.5. This list is then converted in a series of x and y coordinates, where for a given value of the list $l_i$ the corresponding coordinates are $x = i; y = \sum^i_{k=0} l_k$. The coordinates are then normalised such that $(x_0, y_0) = (0, 0)$ and $(x_n, y_n) = (1, 1)$. To convert this series of points to a function and satisfy the constraints mentioned above, we use monotone preserving cubic interpolation \citep{hyman1983accurate}, and normalise the interpolated function with respect to $y$. Our noise generation process has two parameters: the number of points used and the standard deviation of the normal distribution. By setting these parameters accordingly, we generated the two types of noise previously introduced: \emph{short} and \emph{long} (Figure \ref{fig:noise}).

In order to model noise that does not respect the domain knowledge encoded in the mapping function, on the other hand, the only two constraints to be respected were those in Equations \ref{eq:robustness1} and \ref{eq:robustness2}. This type of noise, referred to as \emph{random}, is generated by defining a set of coordinates that were equally spaced on the $x$ axis, and whose $y$ coordinate was a random value between 0 and 1. In order to generate $\mathcal{N}$, we performed standard cubic interpolation and normalised the function between 0 and 1.

In order to isolate the effect of the noise on the progression, in our experimental evaluation, the set of parameters used for the Friction-based progression was not changed.

\paragraph{Experimental results} Our experimental result shows that in the Grid world Maze environment, the \emph{local} and \emph{global} noise have a limited influence on the performance of the agent, as their best performance falls within the confidence interval of the noise-less method. On the other hand, the \emph{random} noise results in lower performance in the initial part of the training and also a lower best performance. It should be noted that the performance was steadily increasing, suggesting that if given more time, this form of noise would converge to a similar value compared to the noise-less method, albeit slower. Moreover, in this domain, the performance of the Friction-based progression with \emph{random} noise is still superior to RCG.

Our experimental result shows that in the MuJoCo maze environments, applying either \emph{local} or \emph{global} noise does not significantly affect the agent's final performance, as all the confidence intervals for different types of noise and the standard algorithm overlap. Figure \ref{fig:noise_fbp} in Appendix \ref{app:a}, in the Point Mass Maze and Directional Point Maze, shows how applying short noise to the progression results in a marginal increase in the average performance in the first half of training. We hypothesise this occurs because the local perturbations caused by the short noise cause the progression to quickly increase the complexity of the environment, which could be beneficial in less complex environments, but detrimental in more complex ones. When applying \emph{random} noise in the same domains, a slight drop in performance can be observed in the Point Maze domain; however, no significant drop in performance can be observed in the directional Point Maze or Ant Maze domains.

In the HFO environment, the best performance achieved by the \emph{local} noise falls within the confidence interval of the best performance of the standard algorithm. However, the deviation between the runs is increased. On the other hand, applying \emph{global} noise results in a drop of $8.8\%$ in the final average probability to score and a noticeable increase in the standard deviation between different runs. When looking at Figure \ref{fig:noise}, the average scoring probability is always higher when no noise is present, although the \emph{local} noise is always within the confidence interval. On the other hand, applying \emph{random} noise has an extremely detrimental effect on this specific domain, resulting in a severe drop in performance. It should be noted that this is the only domain where this phenomenon can be observed, so it is hypothesized that this is due to this being the only domain that does not include multiprocessing.

Finally, in the Predator-prey domain, the performance seems unaffected by either type of noise until $2/3$ of the training time. Both \emph{local} and \emph{global} noise cause an increased confidence interval and a loss of average performance after this point. This effect is more pronounced when adding \emph{global} noise, although the best performance achieved by both types of noise falls out of the confidence interval relative to the noise-less method. Applying \emph{random} noise in this domain results in a slower but steadier increase in performance, with its best value falling within the confidence interval of the other two types of noise.

Our results show that even without changing the parameters of our algorithm, the Friction-based progression is highly resilient to both the \emph{local} and \emph{global} types of noise. Moreover, the \emph{local} noise is found to have a lesser impact on the training compared to the \emph{global} noise. This is an expected result as whilst local perturbations in the complexity of the environment can be addressed by our algorithm, more pronounced perturbations result in the progression functions with a higher interval to not keeping up with the agent's ability to solve the environment. On the other hand, our algorithm is more vulnerable to \emph{random} noise, which is also expected, as this noise results in little correlation between how difficult an MDP is and its complexity. This type of noise results in a negligible drop in performance in 3 domains out of the six, a moderate drop in performance on two domains and a severe drop in performance in one domain. Given that this type of noise essentially randomises half of our framework (the mapping function), these results highlight the resilience of the friction-based progression.

\begin{table}[H]
 \small
 \centering
 \begin{tabular}[t]{ccccccc} 
 \hline
 Variant & GW Maze & Point Maze & Dir. P. M. & Ant Maze & 2v2 HFO & Pred. Prey\\ [0.5ex] 
 \hline
 Standard & \bf94.4 $\pm$ \bf3.9 & \bf99.7 $\pm$ \bf0.4 & \bf97.8 $\pm$ \bf1.2 & \bf95.1 $\pm$ \bf1.4 & \bf75.8 $\pm$ \bf1.4 & \bf249.8 $\pm$ \bf8.5\\
 
 Short noise & \bf92.5 $\pm$ \bf5.4 & \bf100 $\pm$ \bf0.0 & \bf98.6 $\pm$ \bf0.7 & \bf94.3 $\pm$ \bf1.3 & \bf75.2 $\pm$ \bf3.6 & 225.7 $\pm$ 14.0\\
 
 Long noise & \bf94.8 $\pm$ \bf4.3 & \bf99.8 $\pm$ \bf0.2 & \bf98.0 $\pm$ \bf1.0 & \bf94.6 $\pm$ \bf1.0 & 67.0 $\pm$ 7.2 & 218.9 $\pm$ 14.1\\
 
  Random noise & 77.4 $\pm$ 10.8 & 97.9 $\pm$ 1.3 & \bf99.2 $\pm$ \bf0.4 & \bf93.1 $\pm$ \bf1.4 & 21.6 $\pm$ 9.6 & 181.0 $\pm$ 38.0\\
 
 \hline
\end{tabular}\\
\caption{Best performance on each domain. All the methods within a $95\%$ confidence interval from the best performing method are in bold.} \label{tab:noise}
\end{table}

\begin{figure*}[!t]
\centering
   \begin{minipage}{0.43\textwidth}
     \includegraphics[width=\textwidth]{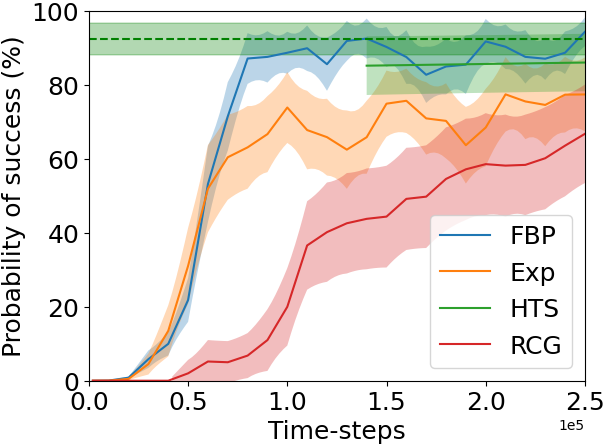}
   \end{minipage}
   \hspace{1cm}
   \begin{minipage}{0.43\textwidth}
     \centering
     \includegraphics[width=\textwidth]{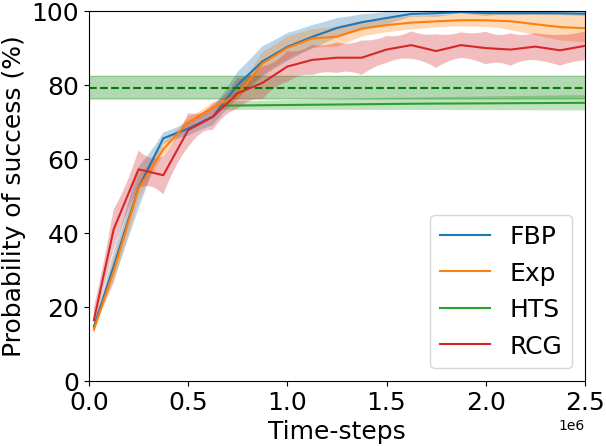}
     
   \end{minipage}
   \begin{minipage}{0.43\textwidth}
     \centering
     \includegraphics[width=\textwidth]{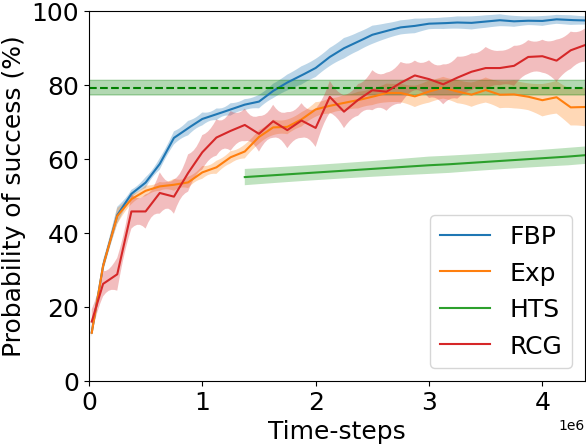}
     
   \end{minipage}
   \hspace{1cm}
   \begin{minipage}{0.43\textwidth}
     \centering
     \includegraphics[width=\textwidth]{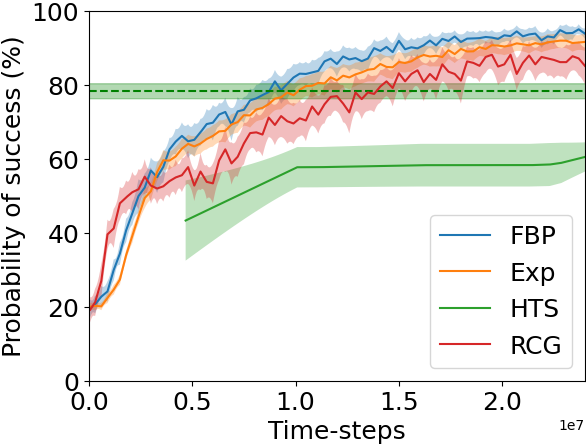}
     
   \end{minipage}
   \begin{minipage}{0.43\textwidth}
     \centering
     \includegraphics[width=\textwidth]{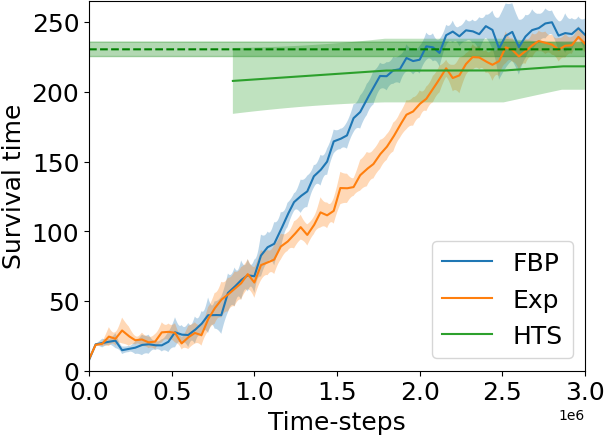}
     
   \end{minipage}
   \hspace{1cm}
   \begin{minipage}{0.43\textwidth}
     \centering
     \includegraphics[width=\textwidth]{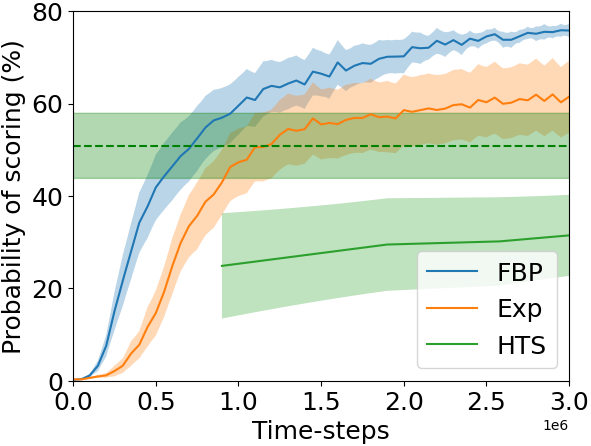}
     
   \end{minipage}
   \caption{Performance of different Progression functions, \emph{Reverse Curriculum} and \emph{HTS-CR} on six test domains, with the dashed line representing the final performance for \emph{HTS-CR}. (From top left to bottom right: Grid World Maze, Point Mass Maze, Directional Point Maze, Ant Maze, Predator-prey, Half Field Offense)} \label{fig:env_res}
\end{figure*}

\subsection{Comparison with the state of the art}

This section seeks to verify whether the approaches derived from our framework can outperform state-of-the-art Curriculum Learning algorithms.

In this experimental evaluation, we compare the Friction-based progression and Exponential progression to two state-of-the-art Curriculum Learning techniques on our testing domains. As the assumptions of Reverse Curriculum Generation are not met by two of our testing domains, as previously discussed, the comparison to this algorithm will take place on the remaining four. The results of this comparison are plotted in Figure,\ref{fig:env_res} and the best performance of each algorithm on each domain is reported in Table \ref{tab:curr}

\paragraph{Experimental results} In the Grid World Maze domain, the Friction-based progression and HTS-CR both converge to similar performance, whereas the best performance achieved by the Exponential progression and RCG falls within the confidence interval from each other. It is worth noting that the difference in the success rate of these two methods is about $10\%$ in favour of the Exponential progression, although RCG did not have enough time to converge, suggesting that if given more time, its final performance would be higher.

In the Point Mass Maze domain, the Friction-based progression is the best performing method, with the Exponential progression's best performance being just outside of the confidence interval. RCG is the third-best performing method, keeping up with the two progression functions in the first third of the training but resulting in a lower best performance. Finally, HTS-CR is the worst performing method in this environment, achieving an average success rate of $79.3\%$ on the target task.

In the Directional Point Maze domain, our results show a clear advantage to using the Friction-based progression over all other methods, resulting in a convergence close to the optimal policy already at three-quarters of the training. On the other hand, RCG produces a curriculum with a much higher deviation between different runs and a significantly slower increase in performance. The Exponential progression is clearly outperformed by RCG in this domain, as whilst in the first three-quarters of the training the two methods perform similarly, RCG's performance keeps improving whereas the Exponential progression's performance falls off slightly. As the exponential progression does not take into account the agent's performance, it will progress to a more complex environment regardless of whether the agent can solve the current one. As a lot of the environments where Curriculum Learning is useful have a sparse reward function, this could potentially result in the agent not experiencing any reward for several epochs, resulting in the deterioration of the agent's policy. Out of the six testing domains, this is the only one where this phenomenon can be clearly identified by looking at the agent's performance over time. It should also be noted that the best performance achieved by the Exponential progression in this domain is also the performance achieved by HTS-CR.

In the Ant maze domain, the Friction-based progression was the best performing method, with the Exponential progression being within its confidence interval for most of the training. Reverse Curriculum Generation resulted in a lower performance throughout the training and an increased deviation between runs. Finally, HTS-CR's final performance was the lowest of the algorithms tested on this domain.

As mentioned in Section \ref{sec:rcg_impl}, RCG was not applicable in both the HFO and Predator-prey domains.

The results in the HFO domain are consistent with our other testing domains, where the Friction-based progression is clearly the best performing algorithm, followed by the exponential progression and finally by HTS-CR. In this domain, the final performance of HTS-CR was within the confidence interval of the best performance of the Exponential progression, although the Exponential progression resulted in a $12\%$ increase in the average probability to score.

Finally, in the Predator-prey domain, the Friction-based progression was the best performing method, although the Exponential progression's highest survival time was within its confidence interval. On the other hand, HTS-CR's final survival time was within the confidence interval of the Exponential progression but not within the confidence interval of the Friction-based progression.

Overall our experimental results show that the Friction-based progression was consistently the method with the highest average performance; moreover, the exponential progression's performance was either on par with other state-of-the-art Curriculum Learning algorithms or greater. This clearly highlights the benefits of using either progression function and also shows the potential of the framework defined in this paper as more Progression functions are created.




\begin{table}[H]
 \small
 \centering
 \begin{tabular}[t]{ccccccc} 
 \hline
 Algorithm & GW Maze & Point Maze & Dir. P. M. & Ant Maze & 2v2 HFO & Pred. Prey\\ [0.5ex] 
 \hline
 F. B. Pr. & \bf94.4 $\pm$ \bf3.9 & \bf99.7 $\pm$ \bf0.4 & \bf97.8 $\pm$ \bf1.2 & \bf95.1 $\pm$ \bf1.4 & \bf75.8 $\pm$ \bf1.4 & \bf249.8 $\pm$ \bf8.5\\
 
 Exp. Pr. & 77.5 $\pm$ 10.1 & 97.5 $\pm$ 1.7 & 79.3 $\pm$ 2.9 & \bf92.0 $\pm$ \bf2.0 & 62.0 $\pm$ 7.9 & \bf239.3 $\pm$ \bf7.9\\
 
 HTS-CR & \bf92.4 $\pm$ \bf4.3 & 79.3 $\pm$ 3.0 & 79.3 $\pm$ 2.0 & 78.3 $\pm$ 2.0 & 50.8 $\pm$ 7.0 & 230.4 $\pm$ 5.2\\

 Rev. Curr. & 66.8 $\pm$ 13.3 & 90.8 $\pm$ 3.7 & 90.8 $\pm$ 4.5 & 88.2 $\pm$ 4.0 & n/a & n/a\\

 
 \hline
\end{tabular}\\
\caption{Best performance on each domain. All the methods within a $95\%$ confidence interval from the best performing method are in bold.} \label{tab:curr}
\end{table}

\section{Conclusion}

We introduced a novel curriculum learning framework that uses Progression Functions and Mapping functions to build an online Curriculum tailored to the agent. The framework  provides a direct way to encode high-level domain knowledge whenever creating a Curriculum via Mapping functions. We also introduced two algorithms within our framework, the Exponential and Friction-based progression functions, which we extensively tested in our comparative evaluation against two state-of-the-art Curriculum Learning techniques on six domains. 
Our experimental results show how using multiprocessing not only increases the speed of our approach but also tends to result in a higher performance; we also demonstrated the resilience of the Friction based progression to noise even without changing its parameters. Finally we showed the benefits of the techniques derived from our framework compared to existing approaches. In particular, the benefits of using the Friction-based progression to generate a Curriculum based on the agent's performance were highlighted by this being the best performing method on all our test domains.

\newpage

\appendix

\section*{Appendix A.} \label{app:a}

This appendix contains the proofs for the theorem in Section \ref{sec:adaprog} and Equation \ref{eq:linexpeq}.

\bigskip


\noindent
{\bf Theorem} {\it Let $\bar{p}_{[x, y]}$ be the average value of the performance function between $t = x$ and $t = y$. Then the Friction-based progression with mass of the object $m$ ends when

\[\bar{p}_{[t+1-i, t]} - \bar{p}_{[1, i]} = \frac{1}{mg}\]

}

\bigskip

\noindent
{\bf Proof}. To find the necessary conditions in order for the Friction-based progression to end we derive an alternative definition of Equation \ref{eq:fbp.sp} where we expand $s_{t-1}$. Firstly, combining Equations \ref{eq:fbp.sp} and \ref{eq:fbp.fr} results in the following equation:

\[ s_t = s_{t - 1} + m * G * \frac{p_{t-i} - p_t}{i}\]

\noindent
For convenience we substitute $k_t = m * G * \frac{p_{t-i} - p_t}{i}$, which results in the following equation:

\[ s_t = s_{t - 1} + k_t\]

\noindent
As mentioned in Section \ref{sec:adaprog}, to avoid big fluctuations in the difficulty of the environment, we start the progression at time $t = i + 1$. This implies that at $t < i + 1$, the object's speed does not change as no progression is taking place, resulting in the object's speed before then being equal to 1. This information can be used to change the equation above to:

\[ s_t = 1 + \sum_{j= i + 1}^t k_j\]

\noindent
Substituting the original value of $k_t$ results in the equation:

\[ s_t = 1 + \sum_{j= i + 1}^t m G \frac{p_{j-i} - p_j}{i}\]

\noindent
This equation can then be rearranged in the following way:

\[ s_t = 1 + \frac{m G}{i} \sum_{j= i + 1}^t p_{j-i} - p_j\]

\[ s_t = 1 + \frac{m G}{i}( \sum_{j = 1}^{t - i} p_{j} - \sum_{j= i + 1}^t p_j )\]

\[ s_t = 1 + \frac{m G}{i}( \sum_{j = 1}^{i} p_{j} + \sum_{j = i + 1}^{t - i} p_{j} - \sum_{j= i + 1}^{t - i} p_j - \sum_{j = t - i + 1 }^t p_j )\]

\[ s_t = 1 + \frac{m G}{i}( \sum_{j = 1}^{i} p_{j} - \sum_{j = t - i + 1 }^t p_j )\]

\[ s_t = 1 + m G( \frac{\sum_{j = 1}^{i} p_{j}}{i} - \frac{\sum_{j = t - i + 1 }^t p_j}{i} )\]

\noindent
As mentioned in the proof statement above, we introduce the notation $\bar{p}_{[x, y]}$ for the average value of the performance function between $t = x$ and $t = y$. We can therefore use this notation on the equation above to simplify it to:

\[ s_t = 1 + m G( \bar{p}_{[1, i]} - \bar{p}_{[t-i+1, t]} )\]

\noindent
By using this alternative formulation of Equation \ref{eq:fbp.sp} we can finally assert that $s_t = 0$ (the progression ends) when:

\[ \bar{p}_{[t-i+1, t]} - \bar{p}_{[1, i]} = \frac{1}{mG} \]

\hfill\BlackBox

\bigskip

\noindent
{\bf Equation \ref{eq:linexpeq}} states that:

\[\lim_{s \to \infty} \Pi_e(t, \{t_e, s\}) = \Pi_l(t, t_e)\]

\noindent
By substituting the equations of the two progression functions and accounting for the fact that $\alpha = \frac{1}{s}$ we get the following equation:

\[\lim_{\alpha \to 0}(min(\frac{\left( e^{\frac{t}{t_e}} \right)^\alpha -1 }{e^{\alpha} -1}, 1)) = min(\frac{t}{t_e}, 1)\]


\noindent
In this equation, the $min$ expressions are used only to ensure that the value returned by the progression functions is always positive, however, if we prove that the left hand side equals to the right hand side without clipping their value, the proof will also be valid after the clipping is added. This consideration results in the equation above becoming:

\[\lim_{\alpha \to 0}(\frac{\left( e^{\frac{t}{t_e}} \right)^\alpha -1 }{e^{\alpha} -1}, 1) = \frac{t}{t_e}\]




\noindent
As the limit converges to the indeterminate form $\frac{0}{0}$, l'Hôpital's rule can be applied:

\[\lim_{\alpha \to 0}(\frac{\frac{t}{t_e} {\left(e^{\frac{t}{t_e}}\right)}^\alpha}{e^\alpha}) = \frac{t}{t_e}\]


\noindent
The above equation converges to:

\[\frac{t}{t_e} = \frac{t}{t_e}\]

\noindent
therefore proving Equation \ref{eq:linexpeq}.

\hfill\BlackBox

\newpage
\section*{Appendix B.}  \label{app:b}

This appendix includes the plots for evaluating the effect of adding noise to the progression, the three formulations of the Friction-based progression and multiprocessing when creating a Curriculum.

\begin{figure*}[!b]
\centering
   \begin{minipage}{0.43\textwidth}
     \includegraphics[width=\textwidth]{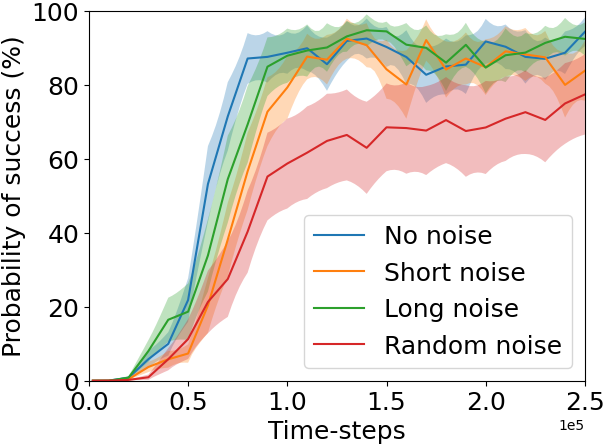}
   \end{minipage}
   \hspace{1cm}
   \begin{minipage}{0.43\textwidth}
     \centering
     \includegraphics[width=\textwidth]{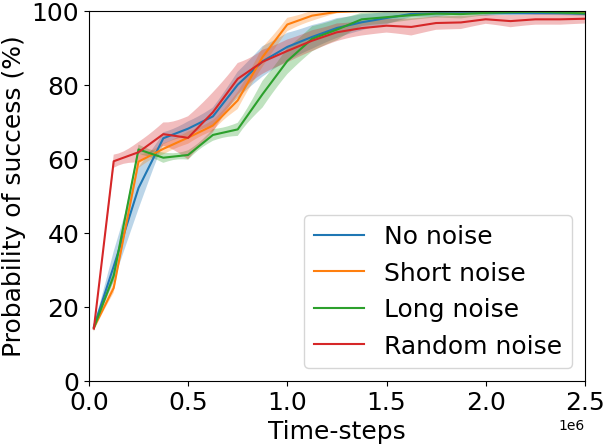}
     
   \end{minipage}
   \begin{minipage}{0.43\textwidth}
     \centering
     \includegraphics[width=\textwidth]{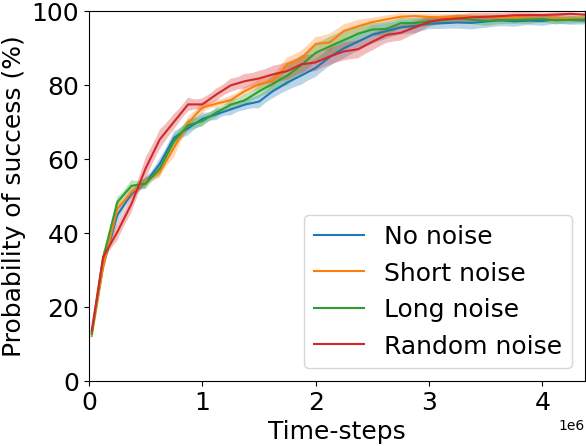}
     
   \end{minipage}
   \hspace{1cm}
   \begin{minipage}{0.43\textwidth}
     \centering
     \includegraphics[width=\textwidth]{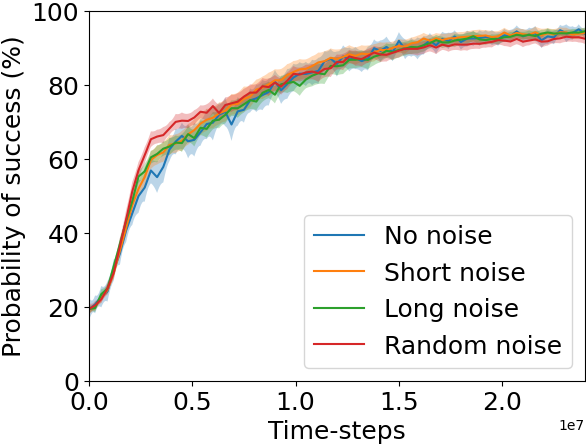}
     
   \end{minipage}
   \begin{minipage}{0.43\textwidth}
     \centering
     \includegraphics[width=\textwidth]{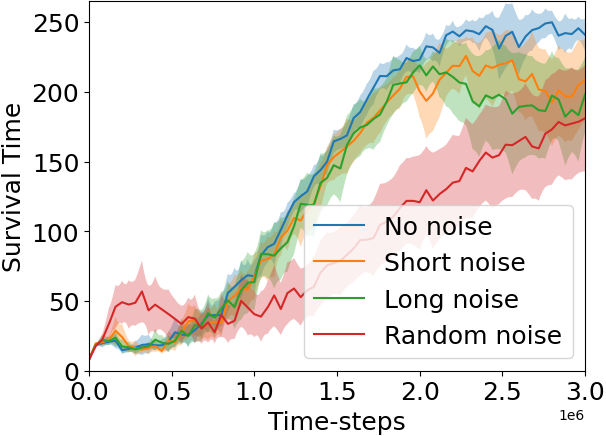}
     
   \end{minipage}
   \hspace{1cm}
   \begin{minipage}{0.43\textwidth}
     \centering
     \includegraphics[width=\textwidth]{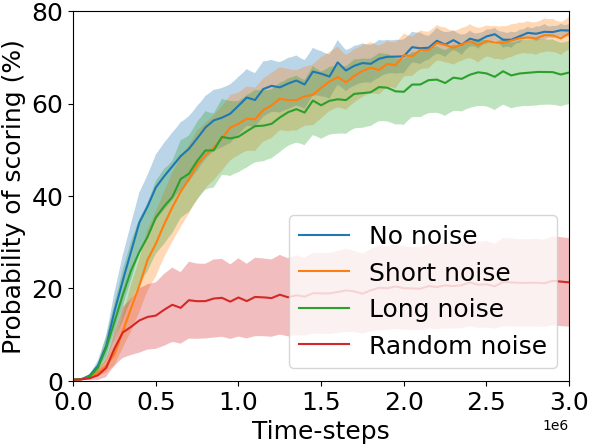}
     
   \end{minipage}
   \caption{Performance of the \emph{Friction-based progression} with different types of noise on six test domains. (From top left to bottom right: Grid World Maze, Point Mass Maze, Directional Point Maze, Ant Maze, Predator-prey, Half Field Offense)} \label{fig:noise_fbp}
\end{figure*}

\begin{figure*}
\centering
   \begin{minipage}{0.43\textwidth}
     \includegraphics[width=\textwidth]{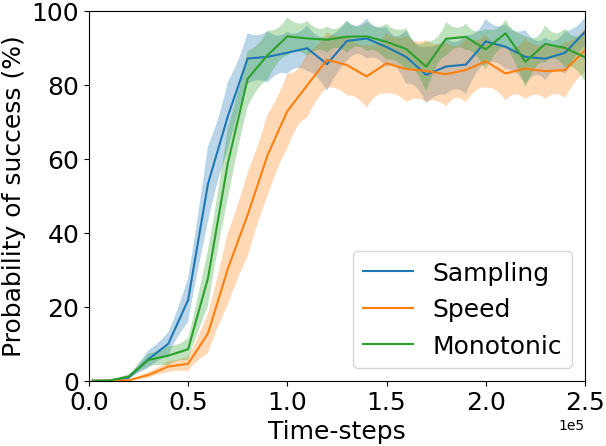}
   \end{minipage}
   \hspace{1cm}
   \begin{minipage}{0.43\textwidth}
     \centering
     \includegraphics[width=\textwidth]{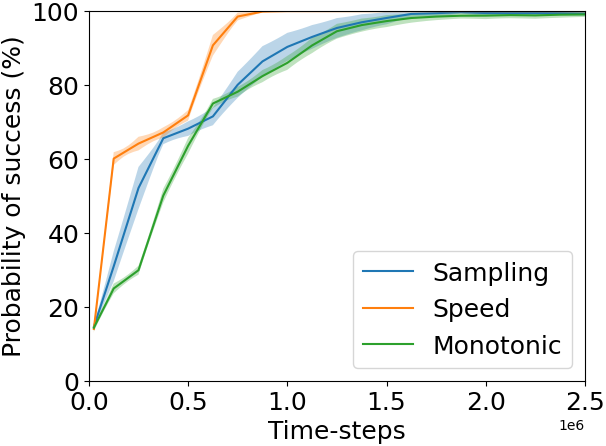}
     
   \end{minipage}
   \begin{minipage}{0.43\textwidth}
     \centering
     \includegraphics[width=\textwidth]{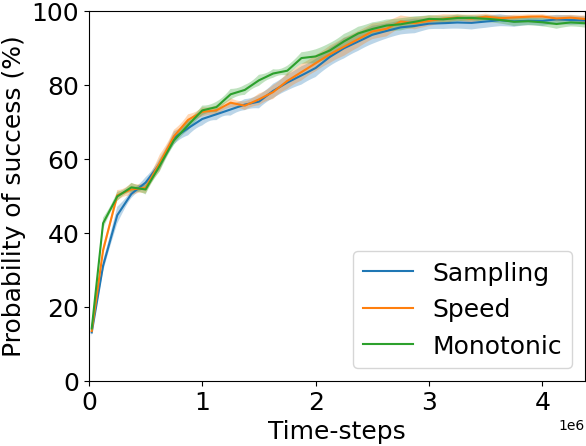}
     
   \end{minipage}
   \hspace{1cm}
   \begin{minipage}{0.43\textwidth}
     \centering
     \includegraphics[width=\textwidth]{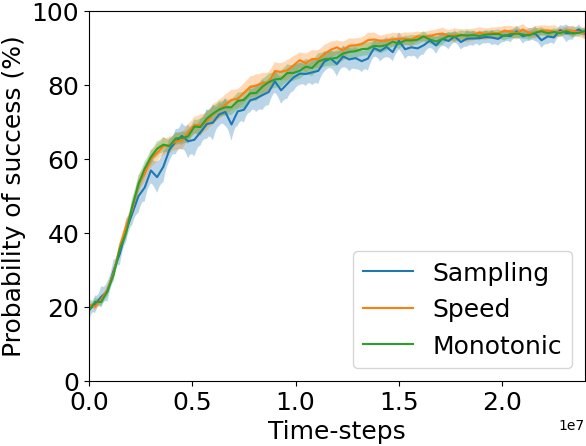}
     
   \end{minipage}
   \begin{minipage}{0.43\textwidth}
     \centering
     \includegraphics[width=\textwidth]{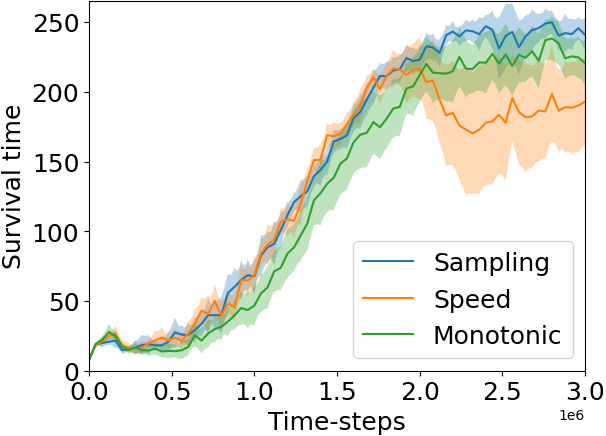}
     
   \end{minipage}
   \hspace{1cm}
   \begin{minipage}{0.43\textwidth}
     \centering
     \includegraphics[width=\textwidth]{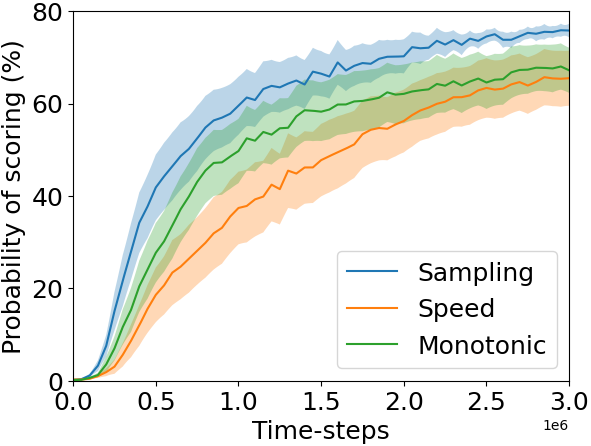}
     
   \end{minipage}
   \caption{Performance of the three formulations of the \emph{Friction-based progression} on six test domains. (From top left to bottom right: Grid World Maze, Point Mass Maze, Directional Point Maze, Ant Maze, Predator-prey, Half Field Offense)} \label{fig:noise_fbp_variants}
\end{figure*}

\begin{figure*}
\centering
   \begin{minipage}{0.43\textwidth}
     \includegraphics[width=\textwidth]{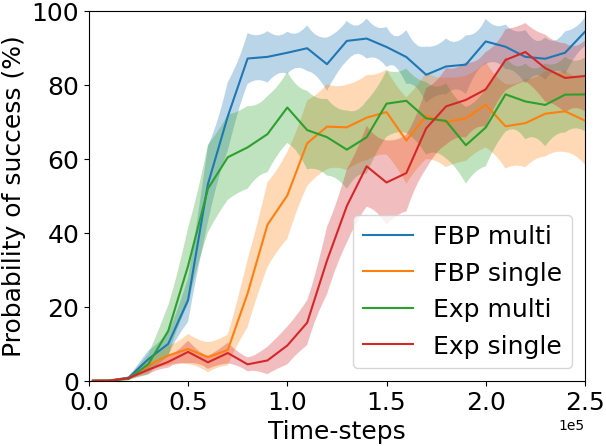}
   \end{minipage}
   \hspace{1cm}
   \begin{minipage}{0.43\textwidth}
     \centering
     \includegraphics[width=\textwidth]{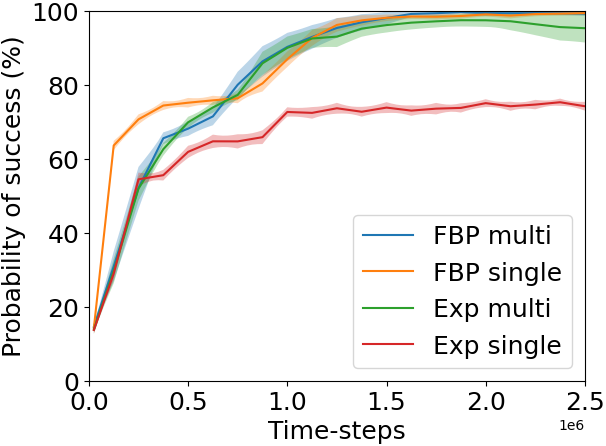}
     
   \end{minipage}
   \begin{minipage}{0.43\textwidth}
     \centering
     \includegraphics[width=\textwidth]{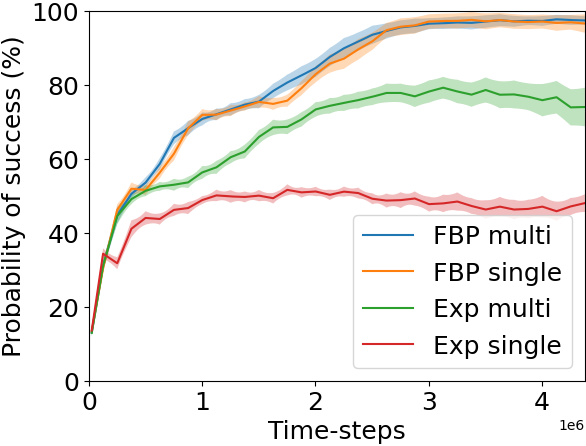}
     
   \end{minipage}
   \hspace{1cm}
   \begin{minipage}{0.43\textwidth}
     \centering
     \includegraphics[width=\textwidth]{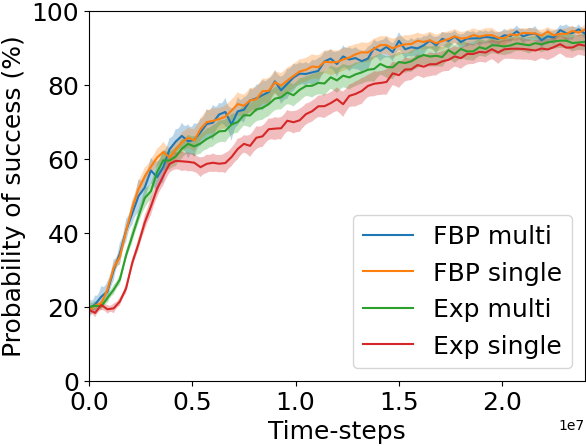}
     
   \end{minipage}
   \begin{minipage}{0.43\textwidth}
     \centering
     \includegraphics[width=\textwidth]{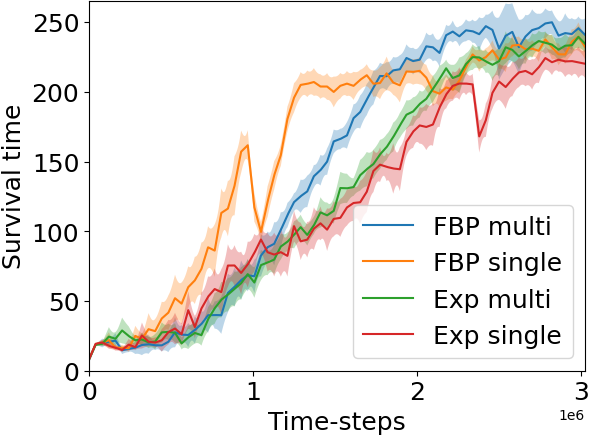}
     
   \end{minipage}
   \hspace{1cm}
   \begin{minipage}{0.43\textwidth}
     \centering
     \hspace{\textwidth}
     
   \end{minipage}
   \caption{Performance of the \emph{Friction-based progression} on five test domains with and without multiprocessing. (From top left to bottom right: Grid World Maze, Point Mass Maze, Directional Point Maze, Ant Maze and Predator-prey)} \label{fig:noise_fbp_mp}
\end{figure*}




















\newpage









\newpage
\bibliography{bibliography}

\end{document}